\documentclass[10pt,twocolumn,letterpaper]{article}

\usepackage[pagenumbers]{cvpr} % To force page numbers, e.g. for an arXiv version

\usepackage{bm}
\usepackage{enumerate}
\usepackage{arydshln}
\usepackage{booktabs}
\usepackage{multirow}
\usepackage{tabularx}
\usepackage{algorithmic}
\usepackage{algorithm}
\usepackage{amsthm}
\usepackage{float}
\usepackage{color}
\usepackage{xfrac}
\usepackage{threeparttable}
\usepackage{siunitx}
\usepackage{caption}
\usepackage{hhline}
\usepackage{array}
\usepackage{threeparttable}
\usepackage{flushend}

\definecolor{cvprblue}{rgb}{0.21,0.49,0.74}
\usepackage[pagebackref,breaklinks,colorlinks,allcolors=cvprblue]{hyperref}

\definecolor{second}{RGB}{255, 255, 200}
\definecolor{best}{RGB}{255, 220, 200}

\def\paperID{6355} % *** Enter the Paper ID here
\def\confName{CVPR}
\def\confYear{2025}

\title{GaussHDR: High Dynamic Range Gaussian Splatting via Learning Unified \\3D and 2D Local Tone Mapping}

\author{
\begin{tabular}[t]{@{}c@{}}
Jinfeng Liu$^1$ \quad Lingtong Kong$^2$ \quad Bo Li$^2$ \quad Dan Xu$^1$
\end{tabular}\\[1ex]
\begin{tabular}[t]{@{}c@{}}
$^1$The Hong Kong University of Science and Technology \quad $^2$vivo Mobile Communication Co., Ltd
\end{tabular}\\[0.5ex]
{\tt\small $\{$jliugk,danxu$\}$@cse.ust.hk, $\{$ltkong,libra$\}$@vivo.com}
}

\begin{document}
\twocolumn[{%
	\renewcommand
	\twocolumn[1][]{#1}%
	\maketitle
	\begin{center}
		\centering        \includegraphics[width=\textwidth]{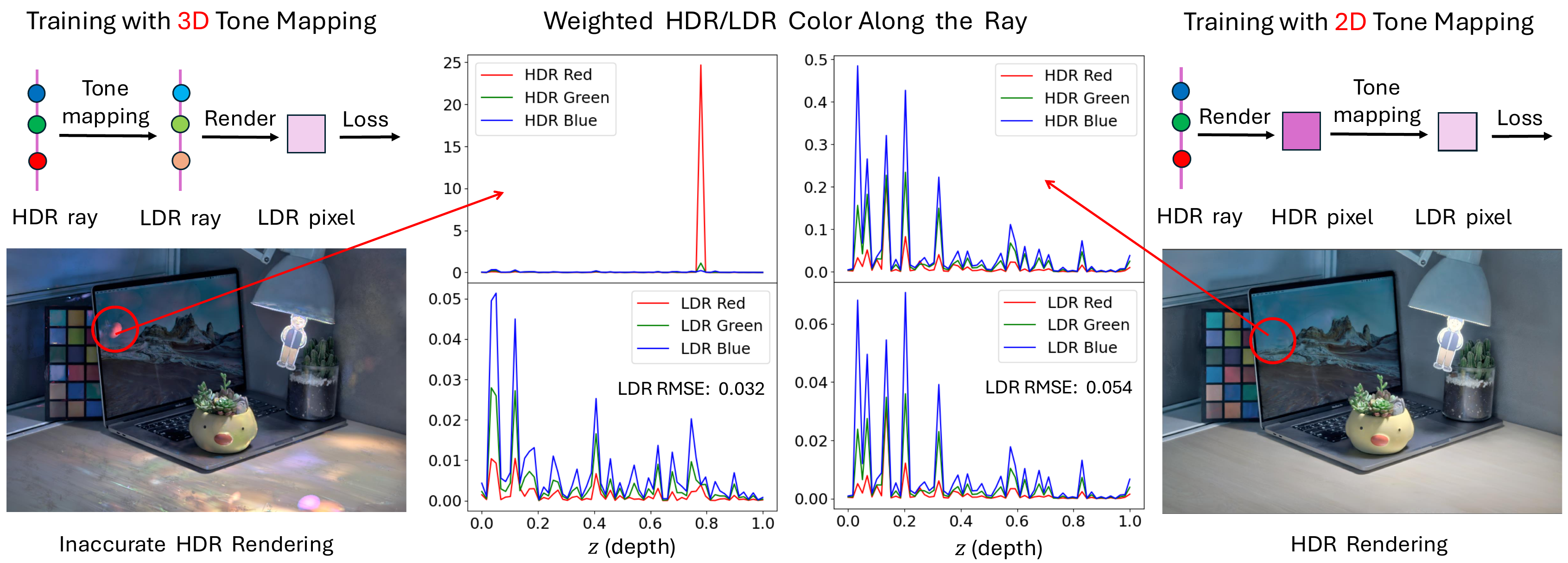}
\vspace{-15pt}
\captionof{figure}{\textbf{3D tone mapping vs. 2D tone mapping}. Training with 3D tone mapping often results in inaccurate HDR rendering, while training with 2D tone mapping degrades LDR rendering quality, leading to a higher LDR RMSE metric compared to 3D tone mapping.}
		\label{fig:teaser}
	\end{center}
}]

\maketitle
\begin{abstract}
High dynamic range (HDR) novel view synthesis (NVS) aims to reconstruct HDR scenes by leveraging multi-view low dynamic range (LDR) images captured at different exposure levels.~Current training paradigms with 3D tone mapping often result in unstable HDR reconstruction, while training with 2D tone mapping reduces the model's capacity to fit LDR images.~Additionally, the global tone mapper used in existing methods can impede the learning of both HDR and LDR representations.
% Moreover, the global tone-mapper employed in existing methods can hinder both HDR and LDR learning.
To address these challenges, we present GaussHDR, which unifies 3D and 2D local tone mapping through 3D Gaussian splatting.
% To address these challenges, we present GaussHDR in this paper, which unifies 3D and 2D local tone-mapping based on 3D Gaussian Splatting. 
Specifically, we design a residual local tone mapper for both 3D and 2D tone mapping that accepts an additional context feature as input.
We then propose combining the dual LDR rendering results from both 3D and 2D local tone mapping at the loss level.
% To conduct 3D tone-mapping, each Gaussian is assigned a new property of context feature. 
% For the 2D tone mapping, each pixel also possesses a context feature rendered from 3D Gaussians. Then, we propose to combine the dual LDR rendering results from 3D and 2D local tone-mapping at the loss level. 
Finally, recognizing that different scenes may exhibit varying balances between the dual results, we introduce uncertainty learning and use the uncertainties for adaptive modulation. Extensive experiments demonstrate that GaussHDR significantly outperforms state-of-the-art methods in both synthetic and real-world scenarios. The project page for this paper is available at \href{https://liujf1226.github.io/GaussHDR}{\textcolor{magenta}{https://liujf1226.github.io/GaussHDR}}.

\end{abstract}

\vspace{-10pt}
\section{Introduction}
\label{sec:intro}
Nowadays, novel view synthesis (NVS) has reached an unprecedented level of advancement with the emergence of Neural Radiance Field (NeRF)~\cite{nerf} and 3D Gaussian Splatting (3DGS)~\cite{3dgs}, enabling realistic reconstruction and real-time rendering. This progress significantly benefits high dynamic range (HDR) NVS, which utilizes multi-view images captured at varying exposure levels to reconstruct HDR scenes. The objective is to render not only novel HDR views but also novel low dynamic range (LDR) views with controllable exposure. Currently, the mainstream paradigm for this task involves extending the color representation from LDR to HDR, and employing a tone mapper to model the camera response function (CRF), which maps the \emph{exposure} (the product of HDR irradiance and camera exposure time) into LDR color. Although methods~\cite{hdr-nerf,hdr-plenoxels,hdr-gs,hdrgs} based on this framework have demonstrated impressive results, the HDR and LDR fitting still faces severe challenges.

Considering the two LDR rendering methods illustrated in~\cref{fig:teaser}, training with 3D tone mapping tends to get stuck in HDR reconstruction while training with 2D tone mapping degrades LDR fitting ability. HDRGS~\cite{hdrgs} also points out the unstable HDR learning but attributes it to the MLP-based tone mapper. Here, we argue that this instability arises because 3D tone mapping separates HDR rendering from LDR supervision, which may lead to inconsistent HDR and LDR distributions along a ray, as depicted in the left part of~\cref{fig:teaser}. Consequently, training with 3D tone mapping is prone to converge to a local optimum where LDR fits well, but HDR fails. Employing 2D tone mapping can alleviate this problem (see the right part of~\cref{fig:teaser}) as LDR supervision is directly related to HDR rendering. However, the LDR results may deteriorate since HDR (from 0 to $+\infty$) ray accumulation in 2D tone mapping is not as robust as standard LDR (from 0 to 1) ray accumulation. Thus, an intuitive idea is to make 3D and 2D tone mapping complement each other, thereby allowing our NVS system to render both HDR and LDR views with high quality. Nonetheless, the joint learning of 3D and 2D tone mapping is still constrained by the modeling capacity of the tone mapper. 

Tone mapper serves as a bridge to connect HDR and LDR representations. The more accurately it can model real-world CRFs, the better HDR and LDR fitting will be. However, existing approaches~\cite{hdr-nerf,hdr-gs,hdrgs} typically apply a \emph{global} tone mapper across an entire scene. This makes a strong assumption that all 3D points or image pixels share the same tone-mapping characteristics, ignoring the fine-grained differences present at various spatial locations. In fact, local operators~\cite{loco1,loco3,lut} in the field of HDR display have shown the ability to enhance visual details by adaptively processing pixels based on their positions.

To tackle the above-discussed problems, we present \textbf{GaussHDR}, a 3DGS-based method for HDR NVS, which unifies the learning of 3D and 2D local tone mapping. We choose 3DGS as the underlying representation because it provides comparable results to NeRFs while significantly improving the training and inference speeds. Building upon 3DGS, we first design a \emph{residual local tone mapper} that is shared across the scene and can adapt to both 3D and 2D tone mapping. Unlike the global tone mapper, it accepts an additional feature vector as input to capture local characteristics. To this end, we assign each 3D Gaussian with a new attribute, \ie, a context feature for 3D local tone mapping. The context features can be rendered from 3D Gaussians to 2D pixels in a manner similar to colors, resulting in a feature map for any given view. Consequently, each pixel will also possess its context feature for 2D local tone mapping. Then, we suggest combining the dual LDR rendering results from 3D and 2D local tone mapping at the loss level to facilitate both HDR and LDR learning. Finally, different scenes may exhibit different balances between the dual results, we therefore introduce uncertainty learning for 3D and 2D local tone mapping and use the uncertainties to adaptively modulate their outcomes. Experiments on synthetic and real-world datasets demonstrate that our GaussHDR significantly surpasses state-of-the-art methods. The main contribution of this paper is summarized as follows:
\begin{itemize}
  \item We present GaussHDR, a novel method that improves HDR Gaussian Splatting by unifying 3D and 2D local tone mapping, where we employ a residual local tone mapper to incorporate local spatial characteristics.
  \item We leverage the context feature assigned to each Gaussian as an additional input condition for the tone mapper in 3D local tone mapping. The context features are rendered concurrently with colors to ensure that each pixel possesses a context feature for 2D local tone mapping.
  \item We combine the dual LDR rendering results from 3D and 2D local tone mapping at the loss level and adaptively modulate them via uncertainty learning.
\end{itemize}

\section{Related Works}
\noindent\textbf{Novel View Synthesis.}~NVS has experienced a significant breakthrough with the introduction of NeRF~\cite{nerf}, which utilizes an MLP to implicitly model both the geometry and appearance of a scene. It leverages differentiable volume rendering~\cite{vr}, enabling training with a set of multi-view images. Follow-up NeRF variants aim to enhance its reconstruction quality~\cite{mipnerf,mipnerf360,refnerf} and efficiency~\cite{ingp,tensorf,trimiprf,zipnerf}. Despite these advancements, training and rendering speeds remain bottlenecks due to the dense sampling required in volume rendering. Recently, 3DGS~\cite{3dgs} has emerged as a superior alternative to NeRF, which explicitly represents a scene by a set of 3D Gaussians. Thanks to the parallel rasterization (or splatting~\cite{ewa}) strategy, 3DGS is much more efficient than NeRFs while maintaining comparable rendering quality. Subsequent works have improved the functionality of 3DGS by incorporating anti-aliasing capabilities~\cite{analytic-splatting,mipsplatting,sa-gs}, applying depth regularization~\cite{dn-gs,dr-gs} and reducing memory overhead~\cite{scaf-gs,hac,lightgs,contextgs}. However, standard NVS methods only work on well-exposed LDR views and lack the capability to recover HDR scenes.

\noindent\textbf{High Dynamic Range Imaging.}~Traditional HDR imaging approaches reconstruct HDR images by merging multi-exposure LDR images captured from a fixed viewpoint~\cite{4392748} or calibrating CRF from these images~\cite{paul}. These methods often encounter motion artifacts when applied to dynamic environments. To mitigate this issue, later techniques propose estimating optical flow to detect motion regions in the LDR images, followed by removal~\cite{katr} or alignment~\cite{nima} during the fusion stage. Recent advancements in deep learning have led to the exploration of using CNNs~\cite{gab,jung,safnet} and Transformers~\cite{rufeng,zhen,jou} to merge multi-exposure LDR images into a single HDR image. Nonetheless, most HDR imaging techniques require ground truth HDR images for supervision and cannot synthesize novel HDR views.

\noindent\textbf{High Dynamic Range Novel View Synthesis.}~Existing HDR NVS methods can be classified into two main groups. The first group directly utilizes noisy RAW data~\cite{9878457,wang2024cinematicgaussiansrealtimehdr,le3d,hdrsplat,chaos}, primarily for nighttime scenes. This paper focuses on the second group, which employs multi-exposure and multi-view LDR images for training~\cite{hdr-nerf,hdr-plenoxels,hdr-gs,hdrgs,10550564}. Among them, HDR-NeRF~\cite{hdr-nerf} first proposes learning HDR radiance fields by predicting the HDR irradiance for each sampled point instead of the original LDR color. It also uses a tone mapper to convert the HDR irradiance into the LDR color. Subsequent methods, such as HDR-Plenoxels~\cite{hdr-plenoxels} and HDR-GS~\cite{hdr-gs}, aim to improve efficiency by replacing the NeRF representation with Plenoxels~\cite{9880358} and 3DGS, respectively. HDRGS~\cite{hdrgs} suggests employing an asymmetric gird-based tone mapper to enhance the reconstruction quality. HDR-HexPlane~\cite{10550564} extends this task to dynamic scenes using HexPlane~\cite{10204912} representation. However, 3D tone mapping in \cite{hdr-nerf,hdr-gs} leads to unsteady HDR reconstruction while 2D tone mapping in \cite{hdr-plenoxels,hdrgs} harms LDR fitting. Besides, all these methods employ a global tone-mapper with a limited modeling capacity. In contrast, we combine 3D and 2D local tone mapping in this paper to facilitate both HDR and LDR learning.
\section{The Proposed GaussHDR}

\begin{figure*}[t]
\centering
%   \fbox{\rule{0pt}{2in} \rule{0.9\linewidth}{0pt}}
\includegraphics[width=1.0\linewidth]{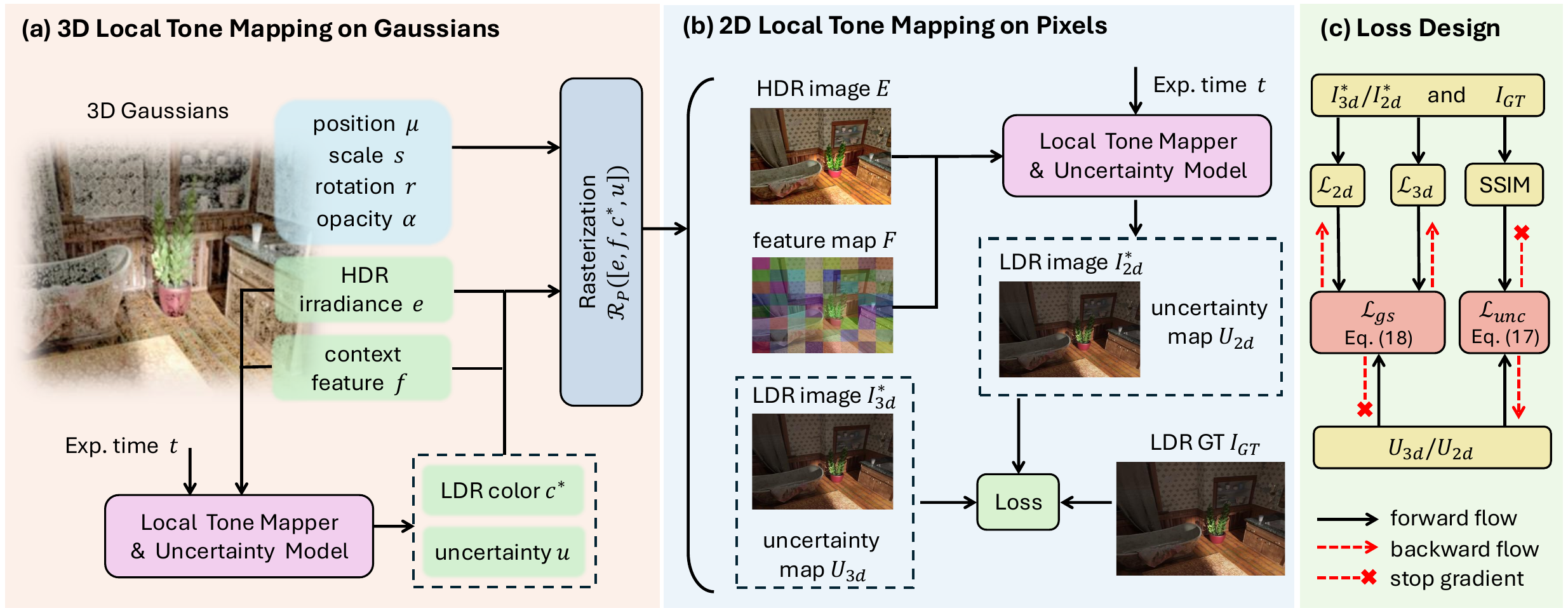}
\caption{\textbf{Overview of the proposed GaussHDR.} (a)  We assign each 3D Gaussian with a context feature for 3D local tone mapping and uncertainty prediction. HDR irradiance, context feature, LDR color and uncertainty are simultaneously rendered onto the image plane. (b) We perform 2D local tone mapping on the rendered HDR image and feature map and predict the uncertainty. (c) We combine the dual LDR rendering results under 3D and 2D local tone mapping at the loss level and utilize their uncertainties for adaptive modulation.} 
\label{fig:framework}
\vspace{-4mm}
\end{figure*}

\subsection{Preliminaries}
\noindent \textbf{Gaussian Splatting.}~3DGS~\cite{3dgs} utilizes a collection of anisotropic 3D neural Gaussians to model the scene. Each Gaussian possesses several attributes, including the position $\mu \in \mathbb{R}^{3}$, rotation quaternion $r \in \mathbb{R}^{4}$, scaling $s \in \mathbb{R}^{3}$, opacity $\alpha \in [0,1]$, and color $c \in \mathbb{R}^{3}$ (represented by spherical harmonics). With these properties, we can represent each Gaussian as $G(x) = e^{-\frac{1}{2}(x-\mu)^T\Sigma^{-1}(x-\mu)}$,
% \begin{equation}
%     G(x) = e^{-\frac{1}{2}(x-\mu)^T\Sigma^{-1}(x-\mu)},
% \end{equation}
where $x$ is an arbitrary 3D position, $\mu$ is Gaussian position (or mean), and $\Sigma$ denotes the covariance matrix. To ensure the positive semi-definite property, $\Sigma$ is decomposed as $\Sigma = RSS^TR^T$, where $R$ is the rotation matrix derived from $r$, and $S$ is the scaling matrix derived from $s$. Each 3D Gaussian $G(x)$ is first transformed into a 2D Gaussian $G'(x)$ on the image plane, as described in~\cite{ewa}. A tile-based rasterizer can then sort the 2D Gaussians and perform $\alpha$-blending:
\begin{equation}
    C(p) = \sum_{i=1}^M c_i \sigma_i \prod_{j=1}^{i-1}(1-\sigma_j),~~~~~  \sigma_i=\alpha_i G'_i(p),
\end{equation}
where $p$ denotes the queried pixel position and $M$ is the number of sorted 2D Gaussians related to the queried pixel.

\noindent \textbf{HDR Radiance Field.}~To adapt the scene representation for HDR NVS, we replace the original LDR color attribute $c \in [0,1]$ with HDR irradiance $e \in [0, +\infty)$. Given the camera exposure time $t$, a tone mapper $\phi$ is employed to model the camera response function (CRF) and map the HDR irradiance to LDR color as $c = \phi(e t)$.
% \begin{equation} \label{eq3}
%     c = \phi(e t).
% \end{equation}
Following previous works~\cite{paul,hdr-nerf,hdr-gs}, we optimize the models in the logarithmic radiance domain to enhance training stability. Specifically, we can rewrite it as $\ln  \phi^{-1} (c)= \ln e + \ln  t$.
% \begin{equation} 
%     \ln  \phi^{-1} (c)= \ln e + \ln  t.
% \end{equation}
Hence, our tone mapper can be transformed into a new function $ g=(\ln  \phi^{-1} )^{-1}$ such that:
\begin{equation} 
    c = g(\ln e + \ln  t),
\end{equation}
where we utilize three different tone mappers to model the RGB channels independently.

\subsection{GaussHDR}
As discussed in \cref{sec:intro}, we aim to seamlessly integrate 3D tone mapping, which demonstrates excellent LDR fitting ability, with 2D tone mapping, which exhibits stable HDR reconstruction capacity, allowing them to complement each other. Besides, we seek to localize the tone mapper to model real-world CRFs more accurately. The key challenges are designing a local tone mapper applicable to both 3D and 2D tone mapping and effectively combining them. We present our solution GaussHDR. An overview is illustrated in \cref{fig:framework}. Firstly, \cref{221} provides a brief formulation of the dual LDR rendering results under 3D and 2D tone mapping. Next, \cref{222} and \cref{223} explain how we localize the tone mapper for 3D and 2D local tone mapping. Finally, \cref{224} introduces the uncertainty-based joint learning of 3D and 2D local tone mapping.

\subsubsection{Dual LDR Renderings} \label{221}
Given neural Gaussians with HDR irradiances $\{e_i\}_{i=1}^N$ and a user-input exposure time $t$, we can transform them into LDR Gaussians $\{c_i\}_{i=1}^N$ through 3D tone mapping:
% \begin{equation}
%     \{c_i\}_{i=1}^N = g(\{\ln (e_i  t)\}_{i=1}^N),
% \end{equation}
\begin{equation}
    c_i = g(\ln (e_i  t)),
\end{equation}
where $g$ is the tone mapper in the logarithmic domain, $i$ means the Gaussian index and $N$ denotes the total number of Gaussians. We then render the LDR image $I_{\text{3d}}$ at a given view from the LDR Gaussians:
\begin{equation} 
    I_{\text{3d}} = \mathcal{R}_P(\{c_i\}_{i=1}^N),
\end{equation}
where $\mathcal{R}_P$ denotes rasterization under camera pose $P$. For simplicity, we omit other auxiliary attributes here. Simultaneously, we can also render the HDR image $E$ at the view from the HDR Gaussians:
\begin{equation} 
    E = \mathcal{R}_P(\{e_i\}_{i=1}^N),
\end{equation}
which can be directly transformed into an LDR image $I_{\text{2d}}$ using 2D tone mapping:
\begin{equation} 
    I_{\text{2d}} = g(\ln(E t)).
\end{equation}
Now we have dual LDR rendering results for each view, $I_{\text{3d}}$ and $I_{\text{2d}}$, under 3D and 2D tone mapping, respectively. Both can be used to calculate loss with ground truth but each has its inherent defects. Although the joint optimization of 3D and 2D tone mapping can combine their strengths, performance is still limited by current global tone mapper.

\subsubsection{3D and 2D Local Tone Mapping}\label{222}
Global tone mapper enforces the same tone-mapping characteristics across the scene, neglecting fine-grained spatial differences. Therefore, we propose to localize the tone mapper for both 3D and 2D tone mapping.

For 3D Gaussians, a direct way could be learning an individual set of tone-mapper parameters for each Gaussian according to its attributes, such as position. Nonetheless, this method is memory-intensive, as each tone mapper contains hundreds of parameters. Therefore, we consider incorporating local characteristics as input to the tone mapper rather than as parameters. For this purpose, MLP is more convenient than grid-based tone mappers~\cite{hdr-plenoxels,hdrgs}. We can still employ a shared MLP across the scene which accepts the value $\ln(e_i t)$ along with some local attributes as input. A naive choice for local attributes could be the Gaussian position (or its Fourier encoding~\cite{nerf}). However, for image pixels, if we similarly use pixel positions as local attributes for tone mapping, we must also account for the distinctions between different views. In fact, HDR-Plenoxels~\cite{hdr-plenoxels} attempts to learn different global tone mappers for various training views, which can only fit existing views without generalization to unseen ones. 

Fortunately, we notice that in language-embedded scene representation~\cite{lerf,legs,langsplat}, the 3D and 2D language feature embeddings are connected through rendering. Pixel features rendered from Gaussian features exist in the same semantic space as Gaussian features. Inspired by this, we assign a new attribute to each Gaussian, \ie, a context feature $f \in \mathbb{R}^{d}$ for local tone mapping, where $d$ is feature dimension. By rendering Gaussian features onto image plane, we can obtain a feature map $F$ at the given view as follows:
\begin{equation} 
    F = \mathcal{R}_P(\{f_i\}_{i=1}^N),
\end{equation}
which indicates that each pixel also corresponds to a context feature for tone mapping. Consequently, we can apply a shared tone mapper $g^*$ to perform 3D local tone mapping:
% \begin{equation} 
%       \{c_i^*\}_{i=1}^N = g^*(\{(\ln (e_i  t), f_i)\}_{i=1}^N),
% \end{equation}
\begin{equation} 
      I_{\text{3d}}^*=\mathcal{R}_P(\{c_i^*\}_{i=1}^N),  ~~~~   c_i^* = g^*(\ln (e_i  t), f_i),
\end{equation}
% \begin{equation} 
%      I_{\text{3d}}^*=\mathcal{R}_P(\{c_i^*\}_{i=1}^N),
% \end{equation}
and 2D local tone mapping:
\begin{equation}  \label{eq12}
    I_{\text{2d}}^* = g^*(\ln(Et), F),
\end{equation}
where $c_i^*$ denotes the intermediate LDR color of the $i$-th Gaussian, $I_{\text{3d}}^*$ and $I_{\text{2d}}^*$ represent the dual LDR rendering results under 3D and 2D local tone mapping, respectively, as illustrated in \cref{fig:framework}(a-b).

\begin{figure}[t]
\centering
%   \fbox{\rule{0pt}{2in} \rule{0.9\linewidth}{0pt}}
\includegraphics[width=1.0\linewidth]{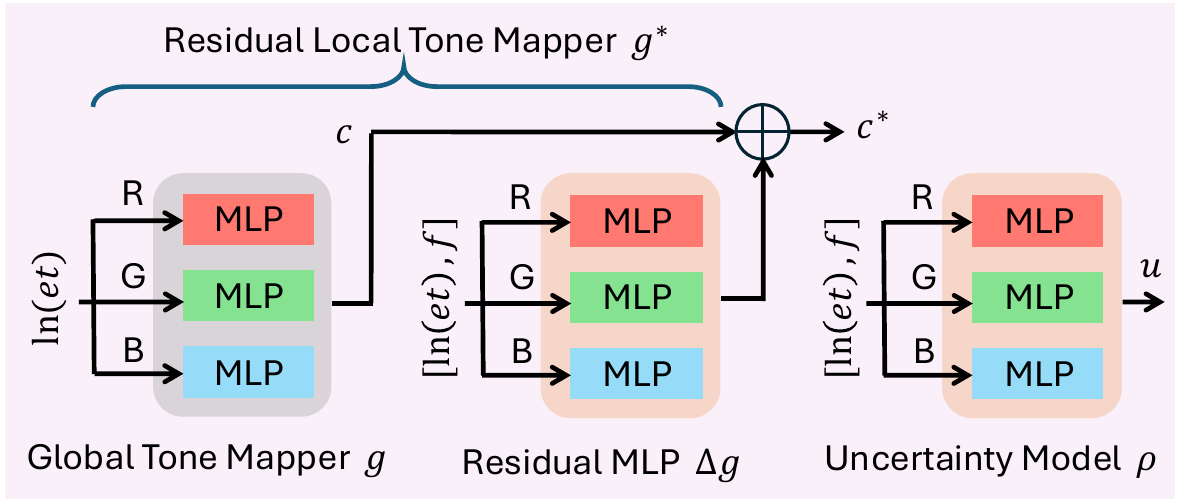}
\vspace{-6mm}
\caption{\textbf{Residual local tone mapper and uncertainty model.} We implement local tone mapper by adding a residual term to the global one. An uncertainty model is utilized to predict the uncertainty of local tone-mapping results.}
\label{fig:ltm}
\vspace{-5.5mm}
\end{figure}

\subsubsection{Residual Local Tone Mapper}\label{223}
In this part, we describe the implementation of our local tone mapper $g^*$. To simplify the learning of $g^*$, we propose modeling it in a residual form. Concretely, we consider the global tone-mapping result as the primary component and add a residual term to incorporate local characteristics, as shown in \cref{fig:ltm}. Consequently, local tone mapping can be expressed as:
\begin{equation} \label{eq13}
    c^* = g^*(\ln(et), f) = g(\ln(et)) + \Delta g([\ln(et), f]),
\end{equation}
where $[~]$ means concatenation, $g$ and $\Delta g$ denote the global tone-mapping MLP and the residual MLP, respectively. In \cref{eq13}, $c^*$ and $e$ correspond to single-channel LDR and HDR colors, as we utilize three independent tone mappers for the RGB channels. The value of $c^*$ is clipped to range $[0,1]$. In this decomposition, the global tone-mapping MLP provides the foundational tonality while the residual MLP captures subtle local variations. By initially developing a well-fitted global tone mapper, we can more easily refine it with local adjustments. In practice, we disable the residual term during the first training stage, training solely with the global tone-mapping results. In the subsequent stage, we enable the residual MLP and optimize $g$ and $\Delta g$ simultaneously using the local tone-mapping results.

\subsubsection{Joint Optimization via Uncertainty}\label{224}
To unify the learning of 3D and 2D local tone mapping, we try to combine the dual LDR rendering results $ I_{\text{3d}}^*$ and $ I_{\text{2d}}^*$, at the loss level. Following 3DGS, we use a combination of DSSIM~\cite{ssim} and L1 losses to compute the image reconstruction error. Formally, given the LDR ground truth $ I_{\text{GT}}$, the loss $\mathcal{L}_{\text{3d}}$ on $I_{\text{3d}}^*$ and loss $\mathcal{L}_{\text{2d}}$ on $I_{\text{2d}}^*$  can be calculated as:
\vspace{-2mm}
\begin{equation}
\begin{aligned}
\mathcal{L}_{\text{3d}} &= \lambda_{\text{d}}\text{DSSIM}(I_{\text{3d}}^*,I_{\text{GT}}) + (1-\lambda_{\text{d}})\|I_{\text{3d}}^*-I_{\text{GT}}\|_1,  \\ \mathcal{L}_{\text{2d}} &= \lambda_{\text{d}}\text{DSSIM}(I_{\text{2d}}^*,I_{\text{GT}}) + (1-\lambda_{\text{d}})\|I_{\text{2d}}^*-I_{\text{GT}}\|_1.
\end{aligned} \vspace{-1mm}
\end{equation}
% \begin{equation} 
%     \mathcal{L}_{\text{3d}}^* = \lambda_{\text{d}}\text{DSSIM}(I_{\text{3d}}^*,I_{\text{GT}}) + (1-\lambda_{\text{d}})\|I_{\text{3d}}^*-I_{\text{GT}}\|_1,
% \end{equation}
% \begin{equation} 
%     \mathcal{L}_{\text{2d}}^* = \lambda_{\text{d}}\text{DSSIM}(I_{\text{2d}}^*,I_{\text{GT}}) + (1-\lambda_{\text{d}})\|I_{\text{2d}}^*-I_{\text{GT}}\|_1.
% \end{equation}
One can simply combine $\mathcal{L}_{\text{3d}}$ and $\mathcal{L}_{\text{2d}}$ through weighted summation. However, we observe that different scenes exhibit varying balances between these losses. To adapt GaussHDR for diverse scenarios, we introduce uncertainty learning~\cite{unc} for both 3D and 2D local tone mapping.

\noindent \textbf{Uncertainty Learning.}~We employ an uncertainty MLP, denoted as $\rho$ in \cref{fig:ltm}, to model the uncertainties of the dual LDR results $I_{\text{3d}}^*$ and $I_{\text{2d}}^*$. This MLP also accepts $\ln(et)$ and the context feature $f$ as input, similar to the local tone mapper. Specifically, we predict the 3D tone-mapping uncertainties $\{u_i\}_{i=1}^{N}$ for all Gaussians and render them onto the image plane to obtain the uncertainty map $U_{\text{3d}}$ for $I_{\text{3d}}^*$, which is formulated as:
\begin{equation} 
     U_{\text{3d}} = \mathcal{R}_P(\{u_i\}_{i=1}^N), ~~~u_i = \rho([\ln(e_i t), f_i]).
\end{equation}
The 2D tone-mapping uncertainty map $U_{\text{2d}}$ for $I_{\text{2d}}^*$ can also be calculated similarly to \cref{eq12}, denoted as:
\begin{equation} 
     U_{\text{2d}} = \rho([\ln(E t), F]).
\end{equation}
Note that we clip the uncertainties with a minimal value of $0.1$~\cite{wild-gs,nerf-on}. Subsequently, we use DSSIM, which causes more stable training dynamics compared to MSE~\cite{nerf-on}, to optimize the uncertainties as follows:
\vspace{-1mm}
\begin{equation} \label{eq14}
\begin{aligned}
\mathcal{L}_{\text{3d-unc}} &= \frac{\text{DSSIM}(I_{\text{3d}}^*,I_{\text{GT}}) }{2U_{\text{3d}}^2} +\lambda_{\text{u}} \ln U_{\text{3d}}, \\
\mathcal{L}_{\text{2d-unc}} &= \frac{\text{DSSIM}(I_{\text{2d}}^*,I_{\text{GT}}) }{2U_{\text{2d}}^2} +\lambda_{\text{u}} \ln U_{\text{2d}}, \\
\mathcal{L}_{\text{unc}} 
 &= \mathcal{L}_{\text{3d-unc}} + \mathcal{L}_{\text{2d-unc}}.
\end{aligned} \vspace{-1mm}
\end{equation}
\noindent \textbf{Joint Optimization.}~Based on the learned uncertainties $U_{\text{3d}}$ and $U_{\text{2d}}$, we can unify 3D and 2D local tone mapping within a joint learning framework, represented as:
\begin{equation} \label{eq17}
\mathcal{L}_{\text{gs}} =\frac{U_{\text{2d}}^2\mathcal{L}_{\text{3d}}+U_{\text{3d}}^2\mathcal{L}_{\text{2d}}}{U_{\text{3d}}^2+U_{\text{2d}}^2},
\end{equation}
where we utilize the uncertainties (or variances) to adaptively modulate $\mathcal{L}_{\text{3d}}$ and $\mathcal{L}_{\text{2d}}$. Following~\cite{nerf-w,nerf-on,wild-gs}, we separate the training of the 3DGS model from uncertainty prediction to ensure that the learned uncertainty is robust. By stopping gradients, as depicted in \cref{fig:framework}(c), $\mathcal{L}_{\text{unc}}$ influences only the uncertainty predictor, while $\mathcal{L}_{\text{gs}} $ affects only the 3DGS model and the tone mapper. 
Besides, for synthetic datasets, we follow HDR-NeRF~\cite{hdr-nerf} to incorporate a unit exposure loss on the global tone mapper $g$ to fix the scale of learned HDR values, enabling HDR quality evaluation, denoted as $\mathcal{L}_{\text{e}}=\|g(0)-0.73\|_2^2$. Therefore, the overall loss function for GaussHDR can be expressed as:
\begin{equation}
\mathcal{L} = \mathcal{L}_{\text{gs}} + \mathcal{L}_{\text{unc}} + \lambda_{\text{e}} \mathcal{L}_{\text{e}}.
\end{equation}
When synthesizing LDR views after training, we can also leverage the uncertainties to merge the dual LDR results:
\begin{equation}
I_{\text{merge}} =\frac{U_{\text{2d}}^2{I}^*_{\text{3d}}+U_{\text{3d}}^2{I}^*_{\text{2d}}}{U_{\text{3d}}^2+U_{\text{2d}}^2},
\end{equation}
where $I_{\text{merge}}$ represents the merged LDR rendering, which is more robust than either of the dual results.

\begin{table*}[t]
  \centering
  \caption{Quantitative comparisons on HDR-NeRF~\cite{hdr-nerf} and HDR-Plenoxels~\cite{hdr-plenoxels} real datasets. Metrics are averaged over all scenes. LDR-OE and LDR-NE denote the LDR results with exposure $\{t_1,t_3,t_5\}$ and $\{t_2,t_4\}$, respectively. The training exposure setting is Exp-3.}
  \vspace{-2mm}
  \resizebox{0.99\textwidth}{!}{
    \begin{tabular}{c|cccccccccccc}
    \toprule
\multirow{3}[4]{*}{Method}  & \multicolumn{6}{c}{HDR-NeRF~\cite{hdr-nerf} Real Scenes}                    & \multicolumn{6}{c}{HDR-Plenoxels~\cite{hdr-plenoxels} Real Scenes} \\
\cmidrule(lr){2-7}\cmidrule(lr){8-13}        
&      \multicolumn{3}{c}{LDR-OE ($t_1,t_3,t_5$)} & \multicolumn{3}{c}{LDR-NE ($t_2,t_4$)} & \multicolumn{3}{c}{LDR-OE ($t_1,t_3,t_5$)} & \multicolumn{3}{c}{LDR-NE ($t_2,t_4$)} \\
\cmidrule(lr){2-4}\cmidrule(lr){5-7}\cmidrule(lr){8-10}\cmidrule(lr){11-13}    
&      PSNR$\uparrow$  & SSIM$\uparrow$  & LPIPS$\downarrow$ & PSNR$\uparrow$  & SSIM$\uparrow$  & LPIPS$\downarrow$ & PSNR$\uparrow$  & SSIM$\uparrow$  & LPIPS$\downarrow$ & PSNR$\uparrow$  & SSIM$\uparrow$  & LPIPS$\downarrow$ \\
\midrule       
HDR-NeRF~\cite{hdr-nerf} $\dag$  & 34.31    &0.954       & 0.062      &    32.29 & 0.949&0.071   & -      &  -     &    -   &   -    &   -    &    -        \\
HDR-GS~\cite{hdr-gs}       & 35.47      &   0.970    &    0.022 &31.66& 0.965   & 0.030  & -      &  -     &    -   &   -    &   -    &    -    \\   
HDR-GS~\cite{hdr-gs} $\dag$ &      35.22     &   0.971    &    0.021   &     31.82  &    0.966   &     0.028  &    31.04   & 0.950      &   0.041    &  25.98     &    0.907   &0.068  \\
\textbf{Ours (3DGS)} &      35.78     &    9.973   &     0.016  &    33.33   &    0.969   &    0.018   & 31.51      & 0.953      &     0.038  &   28.86    &  0.934     &0.046  \\
HDR-Scaffold-GS~\cite{scaf-gs} $*$ &    35.96    &  0.975     &   0.013    &      32.66 &   0.971   &    0.016    &    32.12   &   0.955    &  0.032     &   29.17    &   0.935    & 0.040 \\
\textbf{Ours (Scaffold-GS)} & \textbf{36.77}     &   \textbf{0.977}    &    \textbf{0.011}   &  \textbf{33.92}     &    \textbf{0.974}   &    \textbf{0.014}   &  \textbf{32.85}     &   \textbf{0.959}    & \textbf{0.028}     &   \textbf{29.72}    &  \textbf{0.939}     &\textbf{0.037}        \\    
    \bottomrule
    \end{tabular}}
       \begin{tablenotes}
        \footnotesize
        \item $\dag$ We re-implement HDR-NeRF~\cite{hdr-nerf} and HDR-GS~\cite{hdr-gs} under the same setting (Exp-3) for fair comparison.
        \item $*$ We replace the scene representation in HDR-GS from 3DGS~\cite{3dgs} to Scaffold-GS~\cite{scaf-gs} to establish a baseline for our method utilizing Scaffold-GS.
     \end{tablenotes}
  \label{tab:table1}%
\end{table*}%

\begin{figure*}[t]
\vspace{-2mm}
\centering
%   \fbox{\rule{0pt}{2in} \rule{0.9\linewidth}{0pt}}
\includegraphics[width=0.93\linewidth]{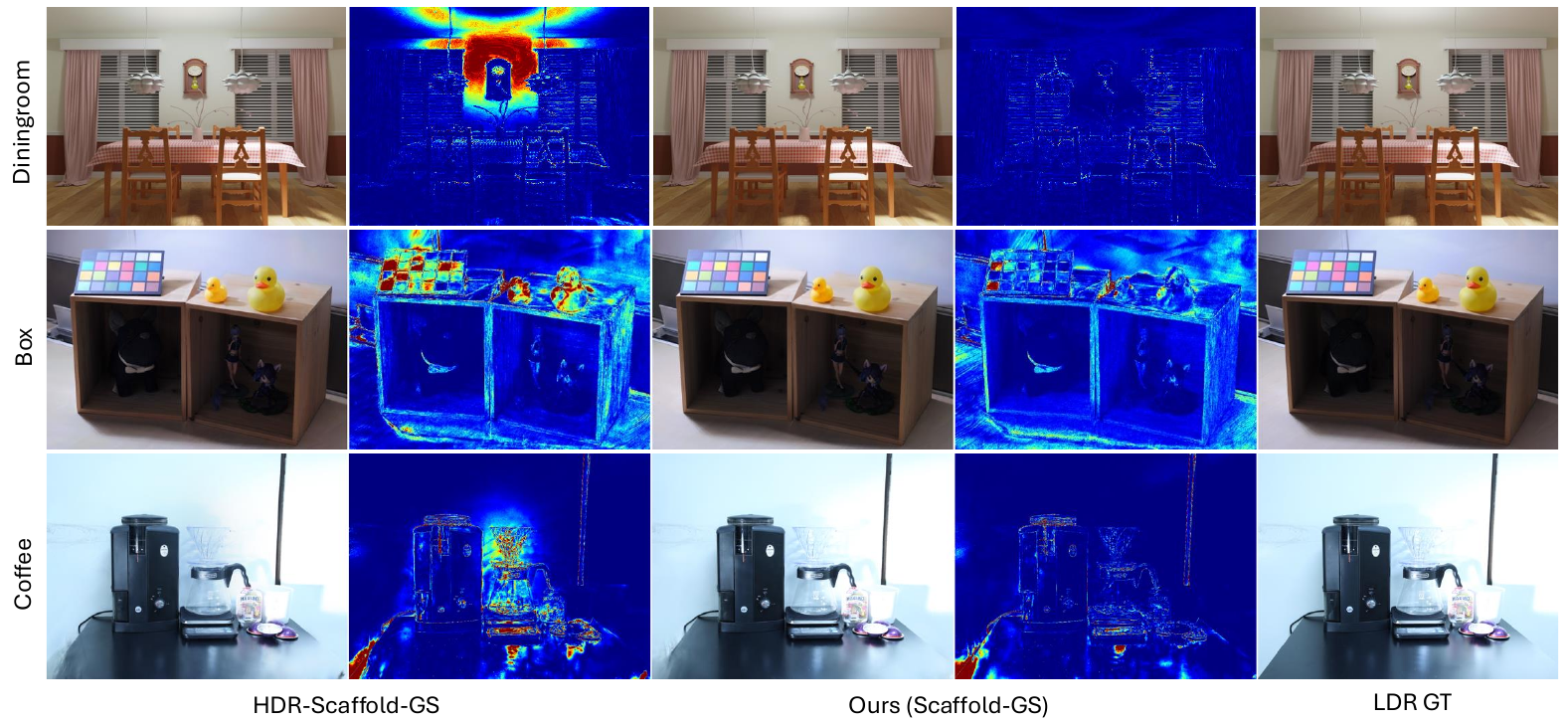}
\vspace{-2mm}
\caption{Qualitative LDR comparisons. Error maps in column 2 and 4 show the MSE error compared to the ground truth, where color from blue to red indicates the error from small to large. Our method can reduce LDR fitting errors in some regions.} 
\label{fig:ldr_vis}
\vspace{-4mm}
\end{figure*}

% You must include your signed IEEE copyright release form when you submit your finished paper.
% We MUST have this form before your paper can be published in the proceedings.

% Please direct any questions to the production editor in charge of these proceedings at the IEEE Computer Society Press:
% \url{https://www.computer.org/about/contact}.
\section{Experiments}

\begin{table*}[t]
  \centering
  \caption{Quantitative comparisons on HDR-NeRF~\cite{hdr-nerf} synthetic datasets. Metrics are averaged over all scenes. LDR-OE and LDR-NE denote the LDR results with exposure $\{t_1,t_3,t_5\}$ and $\{t_2,t_4\}$, respectively. HDR denotes the HDR results. The training setting is Exp-3.}
  \vspace{-2mm}
  \resizebox{0.85\textwidth}{!}{
    \begin{tabular}{c|ccccccccc}
    \toprule
\multirow{2}[4]{*}{Method}    &      \multicolumn{3}{c}{LDR-OE ($t_1,t_3,t_5$)} & \multicolumn{3}{c}{LDR-NE ($t_2,t_4$)} & \multicolumn{3}{c}{HDR}  \\
\cmidrule(lr){2-4}\cmidrule(lr){5-7}\cmidrule(lr){8-10} 
&      PSNR$\uparrow$  & SSIM$\uparrow$  & LPIPS$\downarrow$ & PSNR$\uparrow$  & SSIM$\uparrow$  & LPIPS$\downarrow$ & PSNR$\uparrow$  & SSIM$\uparrow$  & LPIPS$\downarrow$  \\
   
\midrule     
HDR-NeRF~\cite{hdr-nerf} $\dag$    &  40.16   & 0.976 &0.019  &  39.29&0.975&0.020&38.10&0.964&0.020\\
HDR-GS~\cite{hdr-gs}   & 41.10 & 0.982 & 0.011 & 36.33 & 0.977 & 0.016 & 38.31 & 0.972 & 0.013  \\   
HDR-GS~\cite{hdr-gs} $\dag$      &   41.13 &	0.983 &0.009 &39.78 & 0.980& 0.012&	26.98&0.901	&	0.042  \\
\textbf{Ours (3DGS)} &     42.29&	0.985&0.007 & 41.57	&0.985&	0.007  &37.62 & 0.971	&0.016\\
HDR-Scaffold-GS~\cite{scaf-gs} $*$ &    43.16 &	0.989&	0.004 &	42.35 & 0.988&	0.005 & 29.22 &0.950 &0.025\\
\textbf{Ours (Scaffold-GS)} &   \textbf{43.78}   & \textbf{0.990}      &\textbf{0.003}       &    \textbf{43.00}   & \textbf{0.990}      & \textbf{0.004}      &   \textbf{39.02} &\textbf{0.976}&\textbf{0.010}     \\    
    \bottomrule
    \end{tabular}}
       \begin{tablenotes}
        \footnotesize
        \item $\dag$ We re-implement HDR-NeRF~\cite{hdr-nerf} and HDR-GS~\cite{hdr-gs} under the same setting (Exp-3) for fair comparison. Note that the authors of HDR-GS utilize HDR ground truth (GT) for supervision during training on synthetic datasets, whereas our re-implementation does not include this supervision.
        \item $*$ We replace the scene representation in HDR-GS from 3DGS~\cite{3dgs} to Scaffold-GS~\cite{scaf-gs} to establish a baseline for our method utilizing Scaffold-GS.
     \end{tablenotes}
  \label{tab:table2}%
\end{table*}%

\begin{figure*}[t]
\centering
\vspace{-2mm}
%   \fbox{\rule{0pt}{2in} \rule{0.9\linewidth}{0pt}}
\includegraphics[width=0.95\linewidth]{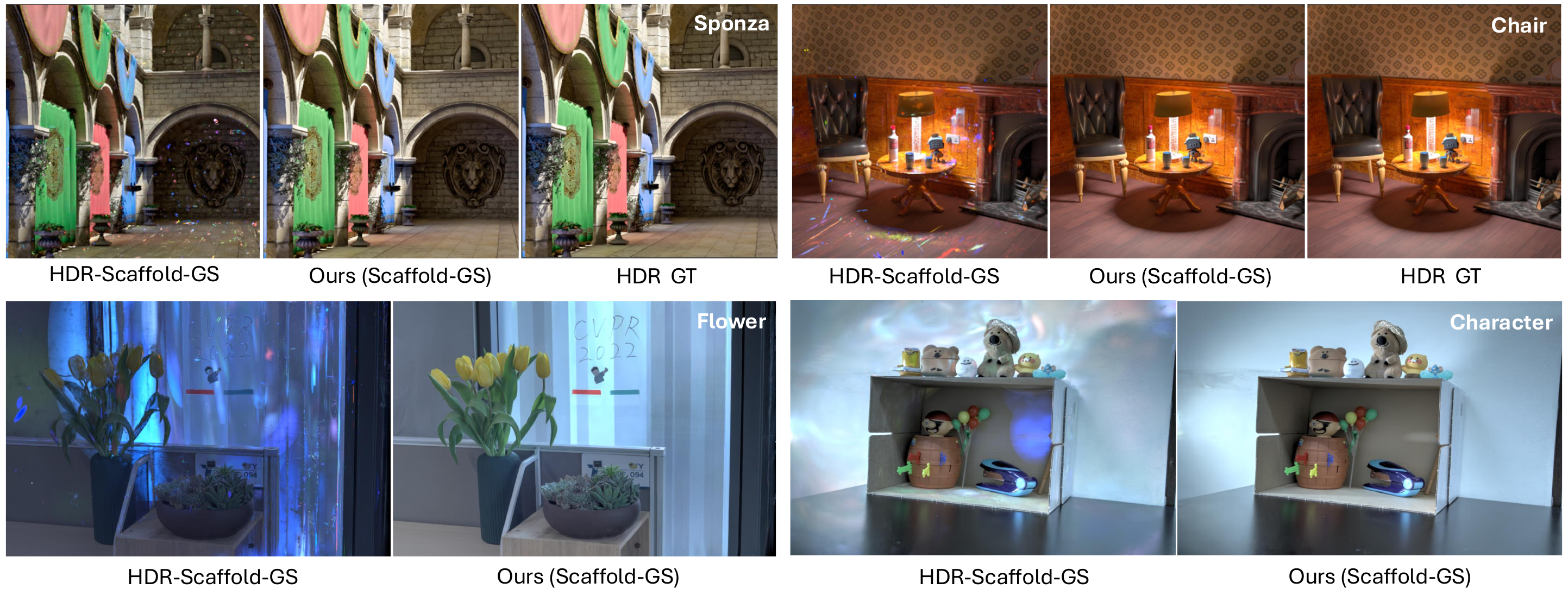}
\vspace{-2mm}
\caption{Qualitative HDR comparisons. Our method leads to stable HDR reconstruction results compared to HDR-Scaffold-GS baseline.} 
\label{fig:hdr_vis}
\vspace{-4mm}
\end{figure*}

\begin{figure}[t]
\centering
% \vspace{-2mm}
%   \fbox{\rule{0pt}{2in} \rule{0.9\linewidth}{0pt}}
\includegraphics[width=0.95\linewidth]{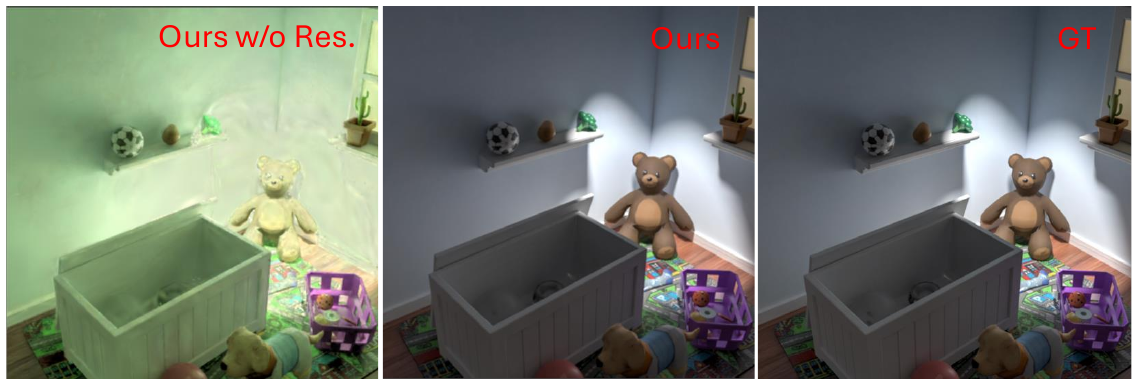}
\vspace{-2mm}
\caption{HDR visualization with residual design or not.} 
\label{fig:res}
\vspace{-5mm}
\end{figure}

\subsection{Experimental Settings}
\noindent \textbf{Datasets.}~We utilize the multi-view and multi-exposure image datasets provided by HDR-NeRF~\cite{hdr-nerf}, including 8 synthetic scenes created in Blender~\cite{blender} and 4 real scenes captured with digital cameras.  Each dataset contains 35 different views, each with 5 distinct exposure times $\{t_1,t_2,t_3,t_4,t_5\}$. The synthetic scenes also include HDR ground truth (GT) for each view. For training, we use images from 18 views with exposure $\{t_1,t_3,t_5\}$. The remaining 17 views, with exposure $\{t_1,t_3,t_5\}$ (LDR-OE) and $\{t_2,t_4\}$ (LDR-NE), along with the HDR GTs of the synthetic datasets, are utilized for evaluation. Additionally, we experiment with another 4 real datasets collected by HDR-Plenoxels~\cite{hdr-plenoxels}, which consist of 40 different views. We use 27 views for training and other 13 views for testing.
\setlength{\parskip}{0pt}

\noindent \textbf{Implementation Details.}~We employ both 3DGS~\cite{3dgs} and its feature-based variant, Scaffold-GS~\cite{scaf-gs}, to verify the effectiveness of the proposed GaussHDR. For Scaffold-GS, since it already provides a feature for each anchor point, we use a linear layer to project this feature into $f \in \mathrm{R}^d$ for local tone mapping, where $d$ is set to 4. The loss weights $\lambda_{\text{d}}$, $\lambda_{\text{u}}$, and  $\lambda_{\text{e}}$ are set to 0.2, 0.5, and 0.5, respectively. All tone-mapper and uncertainty MLPs consist of one hidden layer with 64 channels. Similar to HDR-GS~\cite{hdr-gs}, we recalibrate the camera parameters and extract the initial point clouds for all scenes using COLMAP~\cite{7780814}. Models are trained for 30K iterations using the Adam~\cite{adam} optimizer. During the first 6K iterations, we disable the residual MLP and utilize global tone-mapping results for training. After that, we optimize the whole local tone mapper. All experiments are conducted using Pytorch~\cite{pytorch} on a single RTX 3090 GPU. We experiment with two different training exposure settings. The first (Exp-1) strictly follows HDR-NeRF~\cite{hdr-nerf}, which randomly selects one exposure from $\{t_1,t_3,t_5\}$ for each view and keep it fixed during training. The second (Exp-3) aligns with HDR-GS~\cite{hdr-gs}, where we randomly select an exposure for the sampled view in each training iteration. Exp-1 means only one exposure is used for each view during training, while Exp-3 means all three exposures are accessible during training.

\noindent \textbf{Evaluation Metrics.}~We employ PSNR (higher is better) and SSIM (higher is better) metrics, as well as LPIPS~\cite{lpips} (lower is better) perceptual metric for quantitative evaluation. Following HDR-NeRF~\cite{hdr-nerf}, we quantitatively evaluate HDR results in the tone-mapped domain via the $\mu$-law~\cite{nima} and qualitatively show HDR results tone-mapped by Photomatix pro~\cite{photomatix}.

\begin{table*}[htbp]
  \centering
  \caption{Ablation results on each component of the proposed GaussHDR. TM indicates tone mapping while Res. means the residual form of local tone mapper. All experiments are based on Scaffold-GS under the Exp-1 setting. All the metrics listed here represent PSNR.}
  \vspace{-2mm}
  \resizebox{0.85\textwidth}{!}{
    \begin{tabular}{c|c|c|cccccccccccc}
    \toprule
&\multirow{2}[4]{*}{Method} & \multirow{2}[4]{*}{LDR Test} & \multicolumn{2}{c}{HDR-NeRF~\cite{hdr-nerf} Real Scenes}                    & \multicolumn{3}{c}{HDR-NeRF~\cite{hdr-nerf} Synthetic Scenes} \\
\cmidrule(lr){4-5}\cmidrule(lr){6-8}        
& &       & LDR-OE & LDR-NE  & LDR-OE & LDR-NE & HDR \\
\midrule  
(a)&3D Global TM &  $I_{\text{3d}}$ &33.94 & 31.88& 42.03 & 40.44 & 26.11\\
(b) & 3D Local TM  &  $I_{\text{3d}}^*$ & 34.41 & 32.56 & 42.28 & 41.33 & 26.78\\
(c) & 2D Global TM &  $I_{\text{2d}}$ &32.45 &32.05 & 41.57 & 40.98& 37.96\\
(d) & 2D Local TM &  $I_{\text{2d}}^*$ & 34.19 & 32.92 & 41.95 & 41.34 & 38.35\\
(e)&  3D + 2D Global TM &  $I_{\text{merge}}$ & 33.22 & 32.80 & 41.73 & 41.23 & 38.27\\
(f)&3D + 2D Local TM &  $I_{\text{3d}}^*$ & 35.29 & 33.63 &42.44 & 41.58& \textbf{38.60}\\
(g)&3D + 2D Local TM &  $I_{\text{2d}}^*$ & 35.17 &	33.48 & 42.33 & 41.64& \textbf{38.60}\\
(h)&3D + 2D Local TM (w/o Res.)&  $I_{\text{merge}}$ & 35.08 & 32.63 & 42.78 & 41.73 & 18.46\\
(i)&\textbf{3D + 2D Local TM (Ours)} & $I_{\text{merge}}$ & \textbf{35.37} & \textbf{33.68} & \textbf{42.91} & \textbf{41.97} & \textbf{38.60} \\
    \bottomrule
    \end{tabular}}
    \vspace{-2mm}
  \label{tab:ablation}%
\end{table*}%

\begin{figure*}[t]
\centering
%   \fbox{\rule{0pt}{2in} \rule{0.9\linewidth}{0pt}}
\includegraphics[width=1.0\linewidth]{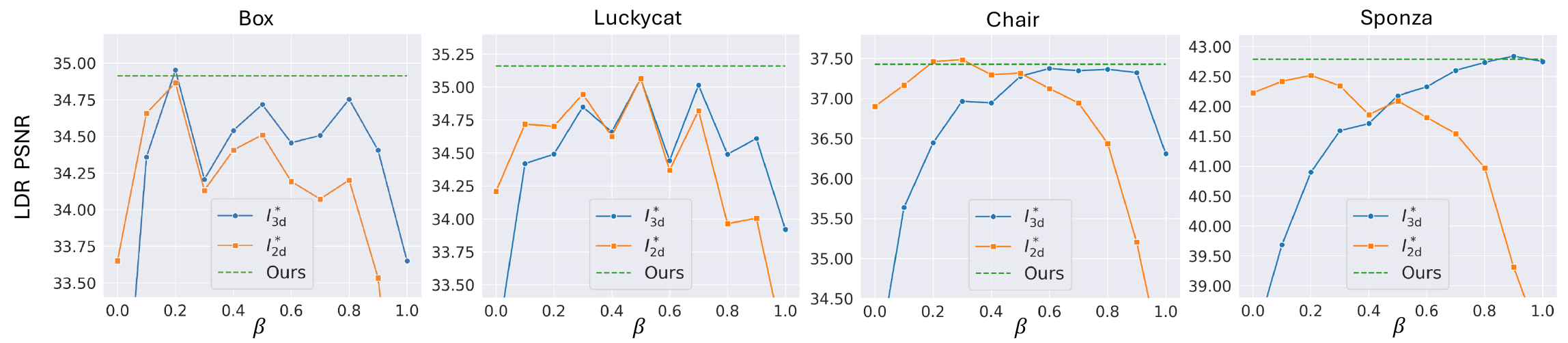}
\vspace{-6mm}
\caption{Performance variations of $I_{\text{3d}}^*$ and $I_{\text{2d}}^*$ with respect to $\beta$ when simply using a loss combination $\mathcal{L}=\beta\mathcal{L}_{\text{3d}}+(1-\beta)\mathcal{L}_{\text{2d}}$. Different scenes exhibit varying optimal values of $\beta$. In contrast, our uncertainty-based modulation (green dash line) can robustly  achieve optimal results across diverse scenes without the selection of hyper-parameter $\beta$. The results of other scenes are in the supplementary materials.} 
\label{fig:ablation}
\vspace{-5mm}
\end{figure*}

\subsection{Performance Comparison}
To establish a baseline for our method utilizing Scaffold-GS, we also conduct experiments with HDR-Scaffold-GS, which replaces the 3DGS representation in HDR-GS~\cite{hdr-gs} with Scaffold-GS. In this part, we mainly compare under the Exp-3 setting. Comparison results under the Exp-1 setting are provided in the supplementary materials.

\noindent \textbf{Quantitative Comparison.}~The quantitative results of real and synthetic datasets are listed in \cref{tab:table1} and \cref{tab:table2}, respectively. It is evident that substituting 3DGS with Scaffold-GS results in improvements across all tracks. This can be attributed to the feature-based GS structure, which can better represent unbounded HDR colors compared to vanilla 3DGS employing spherical harmonics. Furthermore, regardless of whether using 3DGS or Scaffold-GS representations, our method consistently outperforms their corresponding baselines (HDR-GS and HDR-Scaffold-GS, respectively) by a substantial margin. Note that HDR-GS leverages HDR ground truth (GT) for supervision during training on synthetic datasets, resulting in strong HDR performance as shown in \cref{tab:table2}, whereas our implementation dose not include this supervision. Nevertheless, our method, which integrates both 3D and 2D tone mapping, achieves comparable HDR results to HDR-GS. Overall, our GaussHDR demonstrates state-of-the-art performance across all tracks for both real and synthetic datasets.

\noindent \textbf{Qualitative Comparison.} We present visual comparison results for LDR and HDR NVS in \cref{fig:ldr_vis} and \cref{fig:hdr_vis}, respectively. Clearly, our method reduces LDR fitting errors and enhances HDR reconstruction stability compared to the baseline, thereby facilitating both HDR and LDR learning.

\subsection{Ablation Study} 
In this part, we conduct ablation study to dive into each component of GaussHDR. All ablation experiments are based on Scaffold-GS under the Exp-1 setting.

\noindent \textbf{Effect of 3D + 2D Local Tone Mapping.} First, by comparing (a) to (b) and (c) to (d) in \cref{tab:ablation}, we can observe the effectiveness of local tone mapping. No matter whether it is 3D or 2D global tone mapping, the local counterparts can consistently improve the performance. Another observation is that 3D tone mapping demonstrates poorer HDR performance but better LDR-OE fitting ability compared to 2D tone mapping, which has been discussed in \cref{sec:intro}. Furthermore, when unifying the learning of 3D and 2D local tone mapping through uncertainty, both LDR and HDR rendering quality can be further enhanced, as illustrated in \cref{tab:ablation}(i). Additionally, we present the unified learning results under global tone mapping in \cref{tab:ablation}(e), which exhibit inferior performance compared to (i), indicating that local tone mapping is crucial for the joint learning.

\noindent \textbf{Effect of Residual Local Tone Mapper.} To evaluate the effectiveness of the residual design in the local tone mapper, we experiment by utilizing the residual results as the local tone-mapping outcomes. As shown in \cref{tab:ablation}(h), this approach causes a degradation in LDR performance, primarily due to the inherent difficulties in learning a local tone mapper without the benefit of initialization from a global tone mapper. Furthermore, HDR performance also deteriorates significantly, as illustrated in \cref{fig:res}. This deterioration arises owing to the absence of direct HDR constraints from the global tone mapper. Consequently, the local tone mapper heavily relies on the context features, leading to overfitting in LDR and the collapse of HDR.

\noindent \textbf{Effect of Uncertainty Learning.} To investigate the role of uncertainty in our unified learning framework (\cref{eq17}), we perform experiments using a mixed loss function defined as $\mathcal{L}=\beta\mathcal{L}_{\text{3d}}+(1-\beta)\mathcal{L}_{\text{2d}}$, which simply combines $\mathcal{L}_{\text{3d}}$ and $\mathcal{L}_{\text{2d}}$ through weighted summation. The performance variations of $I_{\text{3d}}^*$ and $I_{\text{2d}}^*$ with respect to the hyper-parameter $\beta$ are depicted in \cref{fig:ablation}, revealing that different scenes exhibit varying balances between 3D and 2D tone mapping. Notably,  the uncertainty-based modulation in GaussHDR robustly achieves optimal results across diverse scenes. Moreover, our uncertainty-based merging of the dual LDR results enhances the fusion of their reliable regions, resulting in improved performance compared to either of them, particularly for synthetic scenes, as listed in \cref{tab:ablation}(f,g,i).

\section{Conclusion}
% \vspace{6mm}
In conclusion, we present a novel method, GaussHDR, that enhances HDR Gaussian Splatting by learning unified 3D and 2D local tone mapping. Specifically, we design a residual local tone mapper for both 3D
and 2D tone mapping, which accepts an additional context feature as input. Each Gaussian is assigned a context feature for 3D local tone mapping, and the context features are rendered onto the image plane to ensure that each pixel has an associated context feature for 2D local tone mapping. We combine the dual LDR renderings from 3D and 2D local tone mapping at the loss level and introduce uncertainty learning to adaptively modulate the dual results. 
{
    \small
    \bibliographystyle{ieeenat_fullname}
    \bibliography{main}
}
% \clearpage
% \setcounter{page}{1}
% \begin{figure*}[t]
% \centering
% %   \fbox{\rule{0pt}{2in} \rule{0.9\linewidth}{0pt}}
% \includegraphics[width=0.2\linewidth]{figures/sup.pdf}
% \vspace{-3mm}
% \caption{(a) By multiplying a factor to different RGB channels, we can control the white balance. (b) We visualize the 4-dimension context features channel by channel, which shows that different channels focus on capturing the tone-mapping characteristics of various regions. } 
% \label{fig:sup}
% \end{figure*}

\twocolumn[{%
	\renewcommand
	\twocolumn[1][]{#1}%
	\maketitlesupplementary
	\begin{center}
		\centering        \includegraphics[width=0.9\textwidth]{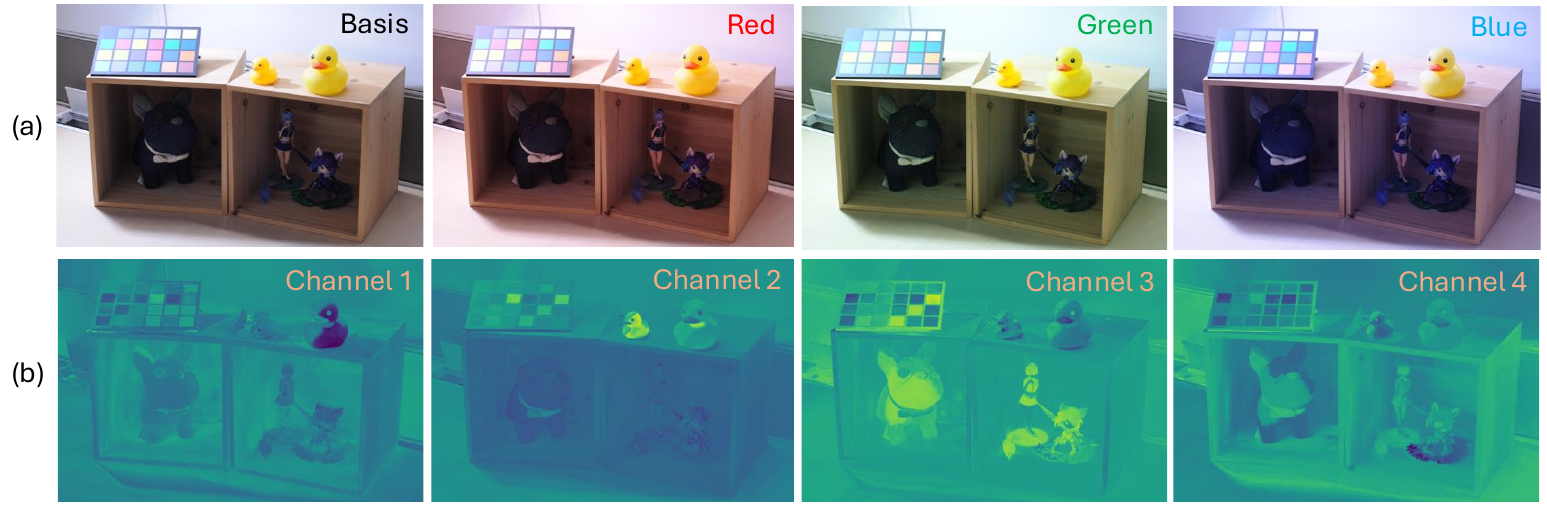}
% \vspace{-15pt}
\captionof{figure}{(a) By multiplying a factor to different RGB channels, we can control the white balance. (b) We visualize the 4-dimension context features channel by channel, which shows that different channels focus on capturing the tone-mapping characteristics of various regions. }
		\label{fig:sup}
	\end{center}
}]

% \maketitlesupplementary

\section{Additional Implementation Details }
\noindent \textbf{Datasets.}~The original image resolutions of HDR-NeRF~\cite{hdr-nerf} synthetic and real datasets are $800\times 800$ and $3216 \times 2136$, respectively. We operate on the $\frac{1}{2}$ scale for synthetic scenes, \ie, a resolution of $400 \times 400$, and the $\frac{1}{4}$ scale for real scenes, \ie, a resolution of $804 \times 534$. For the four additional HDR-Plenoxels~\cite{hdr-plenoxels} real scenes, \emph{desk} and \emph{plant} scenes contain five different exposure times $\{t_1,t_2,t_3,t_4,t_5\}$, while \emph{character} and  \emph{coffee} scenes include only three exposure times  $\{t_1,t_3,t_5\}$. Consequently, we compute the averaged LDR-OE $(t_1,t_3,t_5)$ metrics across the four scenes and the averaged LDR-NE $(t_2,t_4)$ metrics over the \emph{desk} and \emph{plant} scenes. The original resolution of HDR-Plenoxels datasets is $5952 \times 4480$. We operate on the $\frac{1}{6}$ scale, leading to a resolution of $992 \times 746$.

\noindent \textbf{Training Details.}~When employing 3DGS~\cite{3dgs} and Scaffold-GS~\cite{scaf-gs} representations, we adhere to the same training parameters as those specified in \cite{3dgs} and \cite{scaf-gs}, respectively. For our local tone mapper and uncertainty model, the learning rate is initially set to $5\times 10^{-4}$ and exponentially decays to $5\times 10^{-5}$.

\section{Additional Results}
\subsection{Performance Comparison}
The HDR irradiance, tone mapper, and context features are coupled during the scene-specific optimization. However, once we have obtained the HDR radiance field, we can still control the white balance by multiplying a factor to different channels, as shown in \cref{fig:sup}(a). Similarly, we can ignore the context features and switch to other known global tone mappers for different styles, since the context features are only applicable to the learned local tone mapper.

\subsection{Performance Comparison}
\begin{figure*}[t]
\centering
%   \fbox{\rule{0pt}{2in} \rule{0.9\linewidth}{0pt}}
\includegraphics[width=0.96\linewidth]{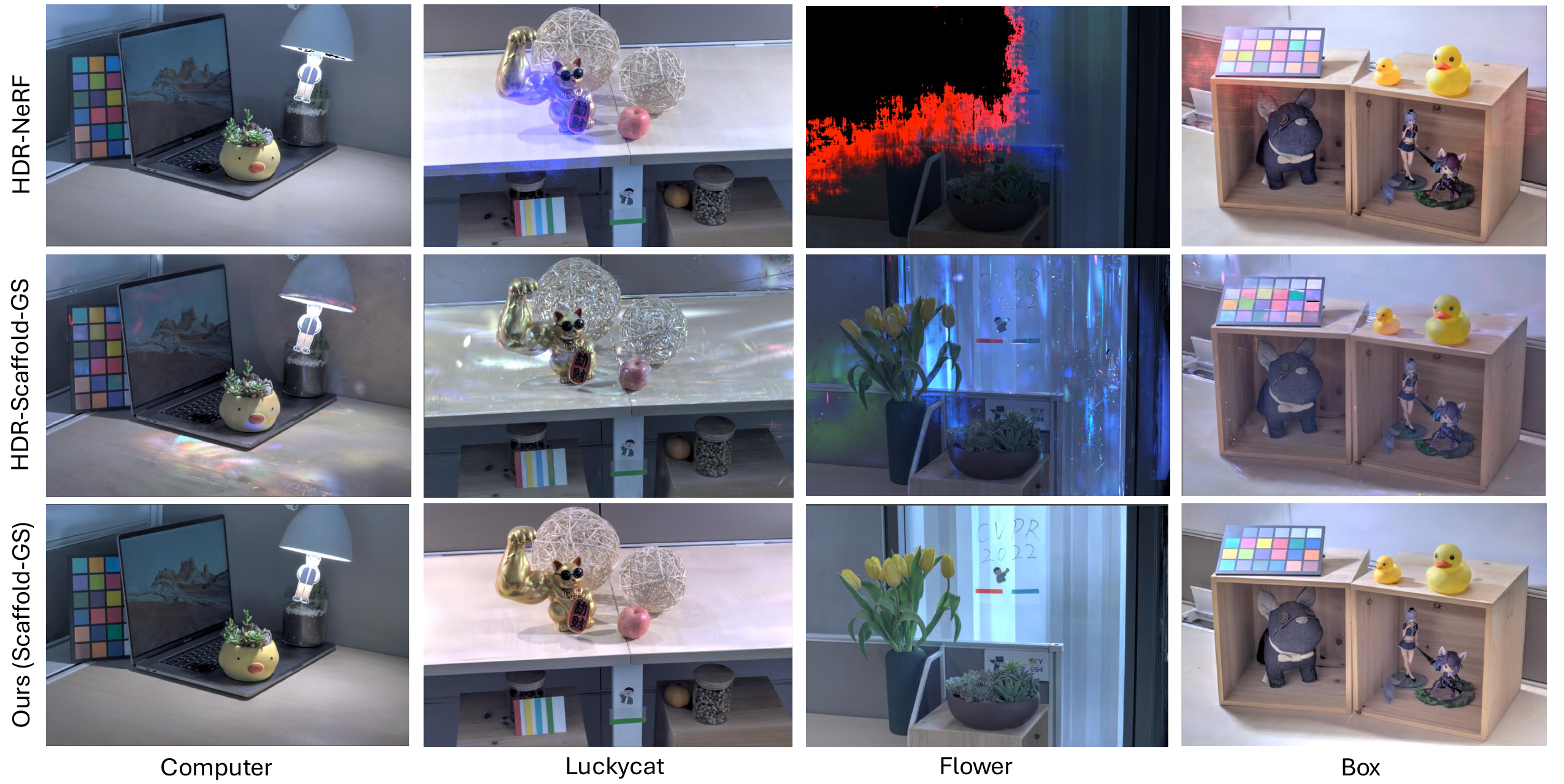}
\vspace{-3mm}
\caption{Qualitative HDR comparisons. Our method leads to stable HDR reconstruction results compared to the baselines.} 
\label{fig:hdr_vis_sup}
\end{figure*}

We have presented the quantitative comparison results under the Exp-3 setting in the main paper. Here, we provide the results under Exp-1 setting, which strictly follows HDR-NeRF~\cite{hdr-nerf} by randomly selecting one exposure from $\{t_1,t_3,t_5\}$ for each view and keeping it fixed during training. The quantitative results for the real and synthetic datasets are listed in \cref{tab:sup_table1} and \cref{tab:sup_table2}, respectively. Per-scene quantitative comparison outcomes are depicted in \cref{tab:hdrnerf-real-4,tab:hdrplen-real-4,tab:hdrnerf-syn-8-1,tab:hdrnerf-syn-8-2}. Additionally, we offer more examples of LDR and HDR qualitative results in the supplementary matearial, as demonstrated in \cref{fig:ldr_vis_sup} and \cref{fig:hdr_vis_sup}, respectively. We also present three video demos corresponding to three different scenes: \emph{bathroom} from HDR-NeRF~\cite{hdr-nerf} synthetic datasets, \emph{flower} from HDR-NeRF real datasets, and \emph{character} from HDR-Plenoxels~\cite{hdr-plenoxels} datasets. All the results further demonstrate that our method achieves state-of-the-art performance by improving both HDR and LDR learning.

\begin{table*}[t]
  \centering
  \caption{Additional ablation results on HDR-NeRF\cite{hdr-nerf} datasets. Pos. indicates using Gaussian position as context feature for local tone-mapping. LDR PSNR denotes the average metric over all 5 exposures. All experiments are under the Exp-1 setting. }
  \vspace{-3mm}
  \resizebox{0.8\textwidth}{!}{
    \begin{tabular}{c|c|cccccccccc}
    \toprule
&\multirow{2}[4]{*}{Method}  & \multicolumn{1}{c}{HDR-NeRF~\cite{hdr-nerf} Real Scenes}                    & \multicolumn{2}{c}{HDR-NeRF~\cite{hdr-nerf} Synthetic Scenes} \\
\cmidrule(lr){3-3}\cmidrule(lr){4-5}       
&      & LDR PSNR & LDR PSNR   & HDR PSNR\\
\midrule  
(a)&Ours-3DGS (with raw Pos.) &32.68 &  40.21  & 36.91\\
(b) & Ours-3DGS (with fourier-encoded Pos.) &32.95& 40.63 & 37.09\\
(c) & Ours-3DGS &\textbf{33.58} &\textbf{41.32}&\textbf{37.41}\\
\midrule  
(d) & Ours-Scaffold-GS (direct learnable weights) & 34.03 & 42.01 & 38.12\\
(e)&   Ours-Scaffold-GS (3-layer MLPs, 128 nodes) & \textbf{34.75} & 42.43 & 38.67\\
(f)&Ours-Scaffold-GS  & 34.69 & \textbf{42.57} & \textbf{38.60}\\
    \bottomrule
    \end{tabular}}
    \vspace{-1mm}
  \label{tab:sup_ablation}%
\end{table*}%

\begin{table*}[t]
  \centering
  \vspace{-2mm}
  \caption{Quantitative comparisons on HDR-NeRF~\cite{hdr-nerf} and HDR-Plenoxels~\cite{hdr-plenoxels} real datasets. Metrics are averaged over all scenes. LDR-OE and LDR-NE denote the LDR results with exposure $\{t_1,t_3,t_5\}$ and $\{t_2,t_4\}$, respectively. The training exposure setting is Exp-1.}
  \vspace{-2mm}
  \resizebox{0.99\textwidth}{!}{
    \begin{tabular}{c|cccccccccccc}
    \toprule
\multirow{3}[4]{*}{Method}  & \multicolumn{6}{c}{HDR-NeRF~\cite{hdr-nerf} Real Scenes}                    & \multicolumn{6}{c}{HDR-Plenoxels~\cite{hdr-plenoxels} Real Scenes} \\
\cmidrule(lr){2-7}\cmidrule(lr){8-13}        
&      \multicolumn{3}{c}{LDR-OE ($t_1,t_3,t_5$)} & \multicolumn{3}{c}{LDR-NE ($t_2,t_4$)} & \multicolumn{3}{c}{LDR-OE ($t_1,t_3,t_5$)} & \multicolumn{3}{c}{LDR-NE ($t_2,t_4$)} \\
\cmidrule(lr){2-4}\cmidrule(lr){5-7}\cmidrule(lr){8-10}\cmidrule(lr){11-13}    
&      PSNR$\uparrow$  & SSIM$\uparrow$  & LPIPS$\downarrow$ & PSNR$\uparrow$  & SSIM$\uparrow$  & LPIPS$\downarrow$ & PSNR$\uparrow$  & SSIM$\uparrow$  & LPIPS$\downarrow$ & PSNR$\uparrow$  & SSIM$\uparrow$  & LPIPS$\downarrow$ \\
\midrule       
HDR-NeRF~\cite{hdr-nerf}    & 31.63
&0.948&0.069&31.43&0.943&0.069& - & -&-&-&-&-  \\
% HDR-NeRF~\cite{hdr-nerf} $\dag$     &   31.95    &   0.951    &     0.066  &    31.75   &    0.950   &    0.067   &       &       &       &       &       &  \\
HDR-GS~\cite{hdr-gs} $\dag$ &  33.56    & 0.964   &  0.024     &   31.18    &  0.960     &    0.027   &   30.18    &    0.943   &   0.044    &    27.63 &  0.916& 0.056    \\
\textbf{Ours (3DGS)} &        34.18    & 0.965      &  0.019     &   32.63    &   0.962    &   0.021    &  30.99     &   0.947    &    0.041   &     28.08&0.923&0.050\\
HDR-Scaffold-GS~\cite{scaf-gs} $*$  & 34.09   &   0.967    &  0.016     &    31.88   &    0.964   &   0.019    &    31.05   &   0.944    &   0.045    &   28.13    &   0.916    &   0.059    \\
\textbf{Ours (Scaffold-GS)}   &  \textbf{35.37} & \textbf{0.972} &  \textbf{0.014} &  \textbf{33.68}  &    \textbf{0.969}   &   \textbf{0.016}   &  \textbf{32.24}    &   \textbf{0.954}    &    \textbf{0.031}   &   \textbf{28.98}    &  \textbf{0.932}     & \textbf{0.041} \\   
    \bottomrule
    \end{tabular}}
       \begin{tablenotes}
        \footnotesize
        \item $\dag$ We re-implement HDR-GS~\cite{hdr-gs} under Exp-1 setting for fair comparison.
        \item $*$ We replace the scene representation in HDR-GS from 3DGS~\cite{3dgs} to Scaffold-GS~\cite{scaf-gs} to establish a baseline for our method utilizing Scaffold-GS.
     \end{tablenotes}
  \label{tab:sup_table1}%
  \vspace{-4mm}
\end{table*}%

\subsection{Ablation on Uncertainty Learning}
\begin{figure*}[t]
\centering
%   \fbox{\rule{0pt}{2in} \rule{0.9\linewidth}{0pt}}
\includegraphics[width=1.0\linewidth]{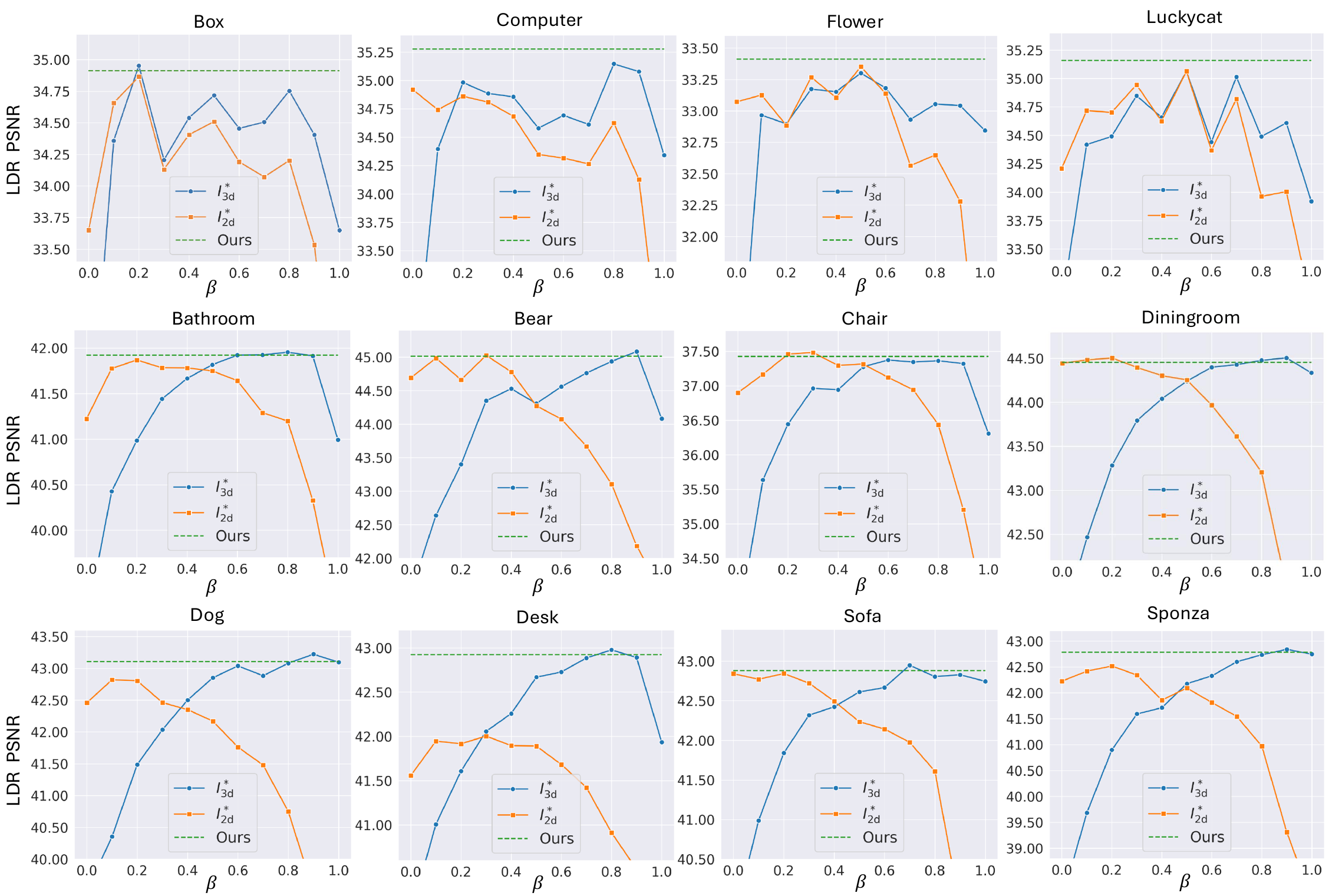}
\vspace{-6mm}
\caption{Performance variations of $I_{\text{3d}}^*$ and $I_{\text{2d}}^*$ with respect to $\beta$ when simply using a loss combination $\mathcal{L}=\beta\mathcal{L}_{\text{3d}}+(1-\beta)\mathcal{L}_{\text{2d}}$. Different scenes exhibit varying optimal values of $\beta$. In contrast, our uncertainty-based modulation (green dash line) can robustly  achieve optimal results across diverse scenes without the selection of hyper-parameter $\beta$. Results include all of HDR-NeRF~\cite{hdr-nerf} real and synthetic scenes.} 
\label{fig:ablation_sup}
\vspace{-5mm}
\end{figure*}

In the main paper, to investigate the effect of uncertainty learning, we perform experiments using a mixed loss function defined as $\mathcal{L}=\beta\mathcal{L}_{\text{3d}}+(1-\beta)\mathcal{L}_{\text{2d}}$, which simply combines $\mathcal{L}_{\text{3d}}$ and $\mathcal{L}_{\text{2d}}$ through weighted summation. Then, we plot the performance variations of $I_{\text{3d}}^*$ and $I_{\text{2d}}^*$ with respect to the hyper-parameter $\beta$. Here, we show the results of more scenes, as depicted in~\cref{fig:ablation_sup}, indicating that different scenes exhibit varying balances between $\mathcal{L}_{\text{3d}}$ and $\mathcal{L}_{\text{2d}}$. However, our uncertainty-based modulation can robustly achieve optimal results across diverse scenes.

\cref{eq14} in the main paper can be viewed as a regularization term that encourages the model to learn meaningful uncertainties or variances. In this part, we also try to directly employ learnable weights for the dual LDR results without any regularization. As shown in \cref{tab:sup_ablation}(d), this approach results in a performance drop, again highlighting the effectiveness of our uncertainty modeling.

\subsection{Ablation on Tone-mapper MLPs}
Since RGB channels may have different tone-mapping characteristics in real-world scenarios. Hence, using channel-specific MLPs for tone mapper is helpful, which has been verified in the HDR-NeRF~\cite{hdr-nerf} paper. Here, we also experiment with larger tone-mapper MLPs with deeper layers and more hidden nodes, as listed in \cref{tab:sup_ablation}(e). We can see that larger MLPs do not bring further improvements, but will increase the training and inference cost.

\subsection{Ablation on Context Feature}
In the main paper, we claim that utilizing pixel positions in image space for 2D local tone mapping is infeasible, as we need to distinguish the same pixel position across different views. However, we can directly leverage Gaussian positions as tone-mapping context features and render them into image space. We conduct ablation experiments for this with 3DGS~\cite{3dgs} representation, since Scaffold-GS~\cite{scaf-gs} utilizes anchor context features to predict Gaussian positions. As listed in \cref{tab:sup_ablation}(a,b), it will cause a degraded performance with either raw Gaussian positions or Fourier-encoded ones. Using Gaussian positions implies that various Gaussians that are far apart must possess distinct tone-mapping characteristics. However, in fact, different spatial locations may exhibit similar characteristics.

We also visualize the 4-dimension context feature channel by channel in \cref{fig:sup}(b), which indicates that different channels focus on capturing the tone-mapping characteristics of various regions. 

\section{Limitations}
Despite the significant enhancements in HDR reconstruction and LDR fitting capabilities provided by GaussHDR, some limitations remain. First, our method relies on COLMAP~\cite{7780814} to extract the initial point cloud and compute camera poses. In other words, we must use tripod-mounted cameras to capture multi-exposure images at each sampled view. However, a more general application scenario could involve utilizing a hand-held camera to capture a monocular video around the scene, where each frame (or view) has a single exposure level, making it difficult to match images captured under different exposures in COLMAP. Therefore, a potential direction for future work is to develop a COLMAP-free version of GaussHDR. Second, it is promising to introduce depth priors to enhance geometry reconstruction by utilizing off-the-shelf depth models~\cite{metric3d,marigold,bdedepth,unidepth,monovifi,depthany}. Finally, our method focuses on static scenes and lacks the ability to perform HDR reconstruction in dynamic environments, which is also an area worth exploring.

\begin{table*}[t]
  \centering
  % \vspace{-360mm}
  \caption{Quantitative comparisons on HDR-NeRF~\cite{hdr-nerf} synthetic datasets. Metrics are averaged over all scenes. LDR-OE and LDR-NE denote the LDR results with exposure $\{t_1,t_3,t_5\}$ and $\{t_2,t_4\}$, respectively. HDR denotes the HDR results. The training setting is Exp-1.}
  \vspace{-2mm}
  \resizebox{0.85\textwidth}{!}{
    \begin{tabular}{c|ccccccccc}
    \toprule
\multirow{2}[4]{*}{Method}    &      \multicolumn{3}{c}{LDR-OE ($t_1,t_3,t_5$)} & \multicolumn{3}{c}{LDR-NE ($t_2,t_4$)} & \multicolumn{3}{c}{HDR}  \\
\cmidrule(lr){2-4}\cmidrule(lr){5-7}\cmidrule(lr){8-10} 
&      PSNR$\uparrow$  & SSIM$\uparrow$  & LPIPS$\downarrow$ & PSNR$\uparrow$  & SSIM$\uparrow$  & LPIPS$\downarrow$ & PSNR$\uparrow$  & SSIM$\uparrow$  & LPIPS$\downarrow$  \\
   
\midrule     
HDR-NeRF~\cite{hdr-nerf}     & 39.07 &0.973& 0.026 &37.53& 0.966 &0.024 &36.40 &0.936&0.018  \\
% HDR-NeRF~\cite{hdr-nerf} $\dag$ &       39.56 & 0.971 & 0.021 & 38.82 & 0.970 & 0.023 & 37.13 & 0.957 & 0.025 \\
HDR-GS~\cite{hdr-gs} $\dag$ &   39.35	&	0.977	&0.012& 38.01& 0.976	&	0.013     &  22.58&0.840	&0.075  \\
\textbf{Ours (3DGS)}  & 41.51	&0.984	&0.008&41.03 &0.983	&0.009 &37.41&0.969&	0.017     \\
HDR-Scaffold-GS~\cite{scaf-gs} $*$ &  42.21 &	0.986	&0.005	&40.44 & 0.986	&0.006  & 26.11 & 0.915&0.062  \\
\textbf{Ours (Scaffold-GS)} & \textbf{42.94}	&\textbf{0.988} &\textbf{0.004}&	\textbf{42.02}& \textbf{0.988}	&	\textbf{0.005} & \textbf{38.60}&\textbf{0.975}	&\textbf{0.011} \\   
    \bottomrule
    \end{tabular}}
       \begin{tablenotes}
        \footnotesize
        \item $\dag$ We re-implement HDR-GS~\cite{hdr-gs} under Exp-1 setting for fair comparison. Note that the authors of HDR-GS utilize HDR ground truth (GT) for supervision during training on synthetic datasets, whereas our re-implementation does not include this supervision.
        \item $*$ We replace the scene representation in HDR-GS from 3DGS~\cite{3dgs} to Scaffold-GS~\cite{scaf-gs} to establish a baseline for our method utilizing Scaffold-GS.
     \end{tablenotes}
  \label{tab:sup_table2}%
\end{table*}%

\begin{table*}[h]
  \centering
  \caption{Per-scene quantitative comparisons on HDR-NeRF~\cite{hdr-nerf} real datasets. LDR-OE and LDR-NE denote the LDR results with exposure $\{t_1,t_3,t_5\}$ and $\{t_2,t_4\}$, respectively. The training exposure setting is Exp-1.}
  \resizebox{0.99\textwidth}{!}{
    \begin{tabular}{c|c|cccccccccccc}
    \toprule
&\multirow{2}[3]{*}{Method}  & \multicolumn{3}{c}{Box} & \multicolumn{3}{c}{Computer} & \multicolumn{3}{c}{Flower} & \multicolumn{3}{c}{Luckycat} \\
\cmidrule(lr){3-5}\cmidrule(lr){6-8}\cmidrule(lr){9-11}\cmidrule(lr){12-14}    
&       & PSNR$\uparrow$  & SSIM$\uparrow$  & LPIPS$\downarrow$ & PSNR$\uparrow$  & SSIM$\uparrow$  & LPIPS$\downarrow$ & PSNR$\uparrow$  & SSIM$\uparrow$  & LPIPS$\downarrow$ & PSNR$\uparrow$  & SSIM$\uparrow$  & LPIPS$\downarrow$ \\
\midrule  
 \multirow{5}[2]{*}{LDR-OE} & HDR-NeRF~\cite{hdr-nerf} & 31.54 & 0.953 & 0.068 & 32.42 &0.950 & 0.077 & 29.81 & 0.948 & 0.069 & 32.85 & 0.938 & 0.062\\
% & HDR-NeRF~\cite{hdr-nerf} $\dag$ & 31.78	& 0.958	& 0.062  & 32.55 &	0.945	& 0.079 & 30.76	& 0.952 &	0.059  & 32.70 &0.948	&0.063   \\
& HDR-GS~\cite{hdr-gs} $\dag$ & 34.39 &0.973&0.016  & 34.41 & 0.967 & 0.020 & 31.62 & 0.956 & 0.037 & 33.82 &0.961&0.022 \\
& \textbf{Ours (3DGS)} &  35.87  &  0.976&0.012 & 34.99 & 0.967 & 0.020 & 31.84 & 0.958 & 0.024 & 34.03 & 0.961 & 0.020\\
& HDR-Scaffold-GS~\cite{scaf-gs} $*$ & 34.66  &  0.974 & 0.012 & 34.55 &0.965&0.017& 32.70 & 0.967&0.019 &34.43&0.963&0.016\\
& \textbf{Ours (Scaffold-GS)} &  \textbf{36.24} & \textbf{0.979} & \textbf{0.010} & \textbf{35.50} & \textbf{0.970} & \textbf{0.015} & \textbf{34.12} & \textbf{0.971} & \textbf{0.014} &\textbf{35.61} &\textbf{0.970}&\textbf{0.015}  \\    
\midrule     
 \multirow{5}[2]{*}{LDR-NE} & HDR-NeRF~\cite{hdr-nerf} &31.40&
0.944 & 0.079  & 31.21 & 0.931 & 0.098 & 30.05 & 0.949 &0.058 &
33.13 & 0.948 & 0.051  \\
% & HDR-NeRF~\cite{hdr-nerf} $\dag$ & 31.25 &	0.957 & 0.062  & 32.61	& 0.947&	0.081  & 30.45	&0.951&	0.062 &32.70	& 0.944&	0.062   \\
& HDR-GS~\cite{hdr-gs} $\dag$ & 30.17 & 0.967 & 0.021 &32.78 & 0.966 & 0.023 & 30.21 & 0.954 & 0.039 & 31.55 &0.954 & 0.026\\
& \textbf{Ours (3DGS)} &  32.57  &0.972& 0.014 & 34.18 &0.966 & 0.022 & 30.52 & 0.956 & 0.027 & 33.24 & 0.956 & 0.022\\
& HDR-Scaffold-GS~\cite{scaf-gs} $*$ & 31.87  &   0.970 & 0.015&33.16&0.964&0.019 &30.68 & 0.964 &0.021&31.80&0.957&0.019 \\
& \textbf{Ours (Scaffold-GS)} & \textbf{32.93} & \textbf{0.974} & \textbf{0.013}  & \textbf{34.95} & \textbf{0.968} & \textbf{0.017} & \textbf{32.35} & \textbf{0.970} & \textbf{0.016} &\textbf{34.48}&\textbf{0.965}&\textbf{0.016}  \\    
    \bottomrule
    \end{tabular}}
       \begin{tablenotes}
        \footnotesize
        \item $\dag$ We re-implement HDR-GS~\cite{hdr-gs} under Exp-1 setting for fair comparison.
        \item $*$ We replace the scene representation in HDR-GS from 3DGS~\cite{3dgs} to Scaffold-GS~\cite{scaf-gs} to establish a baseline for our method utilizing Scaffold-GS.
     \end{tablenotes}
  \label{tab:hdrnerf-real-4}%
\end{table*}%

\begin{table*}[htbp]
  \centering
  \caption{Per-scene quantitative comparisons on HDR-Plenoxels~\cite{hdr-plenoxels} real datasets. LDR-OE and LDR-NE denote the LDR results with exposure $\{t_1,t_3,t_5\}$ and $\{t_2,t_4\}$, respectively. The training exposure setting is Exp-1.}
  \resizebox{0.99\textwidth}{!}{
    \begin{tabular}{c|c|cccccccccccc}
    \toprule
&\multirow{2}[3]{*}{Method}  & \multicolumn{3}{c}{Character} & \multicolumn{3}{c}{Coffee} & \multicolumn{3}{c}{Desk} & \multicolumn{3}{c}{Plant} \\
\cmidrule(lr){3-5}\cmidrule(lr){6-8}\cmidrule(lr){9-11}\cmidrule(lr){12-14}    
&       & PSNR$\uparrow$  & SSIM$\uparrow$  & LPIPS$\downarrow$ & PSNR$\uparrow$  & SSIM$\uparrow$  & LPIPS$\downarrow$ & PSNR$\uparrow$  & SSIM$\uparrow$  & LPIPS$\downarrow$ & PSNR$\uparrow$  & SSIM$\uparrow$  & LPIPS$\downarrow$ \\
\midrule  
 \multirow{4}[1]{*}{LDR-OE} & HDR-GS~\cite{hdr-gs} $\dag$ & 34.96 & 0.978 &0.027 &27.79 & 0.943 &  0.049 & 28.78 & 0.922 &0.045& 29.18 & 0.930 & 0.054\\
& \textbf{Ours (3DGS)} &    36.73   & 0.980 & 0.025 & 28.35 & 0.947 & 0.046 &29.14 & 0.925 & 0.041 & 29.76 & 0.934 & 0.051 \\
& HDR-Scaffold-GS~\cite{scaf-gs} $*$ & 36.81  &0.979&0.029 & 28.31 & 0.946 & 0.043 & 28.64 & 0.919 & 0.049 & 30.44 & 0.930 & 0.059 \\
& \textbf{Ours (Scaffold-GS)} &  \textbf{38.24} & \textbf{0.983} & \textbf{0.018} &\textbf{29.31} & \textbf{0.955} & \textbf{0.032} & \textbf{29.91} & \textbf{0.933} & \textbf{0.035} & \textbf{31.48} & \textbf{0.945} & \textbf{0.040}   \\    
\midrule     
 \multirow{4}[1]{*}{LDR-NE} & HDR-GS~\cite{hdr-gs} $\dag$ & - & - &- &- &- &- & 27.08 &0.912 & 0.050 & 28.17 & 0.920 & 0.061  \\
& \textbf{Ours (3DGS)} &  - & - &- &- &- &- & 27.30 & 0.917 & 0.044 & 28.85 & 0.929 & 0.055 \\
& HDR-Scaffold-GS~\cite{scaf-gs} $*$ & - & - &- &- &- &-  & 27.03 & 0.911 & 0.053 & 29.23 & 0.920 &  0.065    \\
& \textbf{Ours (Scaffold-GS)} & - & - &- &- &- &-  &  \textbf{27.41} & \textbf{0.922} & \textbf{0.038}  & \textbf{30.55} & \textbf{0.941} & \textbf{0.044}      \\  
    \bottomrule
    \end{tabular}}
       \begin{tablenotes}
        \footnotesize
        \item $\dag$ We re-implement HDR-GS~\cite{hdr-gs} under Exp-1 setting for fair comparison.
        \item $*$ We replace the scene representation in HDR-GS from 3DGS~\cite{3dgs} to Scaffold-GS~\cite{scaf-gs} to establish a baseline for our method utilizing Scaffold-GS.
     \end{tablenotes}
  \label{tab:hdrplen-real-4}%
\end{table*}%

\begin{table*}[htbp]
  \centering
  \caption{Per-scene quantitative comparisons on HDR-NeRF~\cite{hdr-nerf} synthetic datasets (Part 1). LDR-OE and LDR-NE denote the LDR results with exposure $\{t_1,t_3,t_5\}$ and $\{t_2,t_4\}$, respectively. HDR denotes the HDR results. The training exposure setting is Exp-1.}
  \resizebox{0.99\textwidth}{!}{
    \begin{tabular}{c|c|cccccccccccc}
    \toprule
&\multirow{2}[3]{*}{Method}  & \multicolumn{3}{c}{Bathroom} & \multicolumn{3}{c}{Bear} & \multicolumn{3}{c}{Chair} & \multicolumn{3}{c}{Diningroom} \\
\cmidrule(lr){3-5}\cmidrule(lr){6-8}\cmidrule(lr){9-11}\cmidrule(lr){12-14}    
&       & PSNR$\uparrow$  & SSIM$\uparrow$  & LPIPS$\downarrow$ & PSNR$\uparrow$  & SSIM$\uparrow$  & LPIPS$\downarrow$ & PSNR$\uparrow$  & SSIM$\uparrow$  & LPIPS$\downarrow$ & PSNR$\uparrow$  & SSIM$\uparrow$  & LPIPS$\downarrow$ \\
\midrule  
 \multirow{5}[2]{*}{LDR-OE} & HDR-NeRF~\cite{hdr-nerf} &36.26 &0.949 &0.037  & 42.91 &0.990 & 0.010 &32.45&0.905&0.081&41.23&0.986&0.010   \\
% & HDR-NeRF~\cite{hdr-nerf} $\dag$ &36.83  & 0.944 & 0.037  &   43.13& 0.990& 0.008 &  34.63& 0.933& 0.058&43.47& 0.988& 0.008   \\
& HDR-GS~\cite{hdr-gs} $\dag$ &   38.06     & 0.963      &   0.020  &41.61	&0.989	&0.005  &35.07	&0.952&	0.024 & 38.26&	0.979&	0.014 \\
& \textbf{Ours (3DGS)} & 41.12& 0.975&	0.008&44.44 &	0.992&	0.003 & 37.05 &	0.968&	0.014 &39.63&0.981	&0.017  \\
& HDR-Scaffold-GS~\cite{scaf-gs} $*$ &  41.21 &	0.977&	0.008 &44.56&	0.992&	0.003 &36.36 &	0.966&	0.016 &44.87&	\textbf{0.994}	&\textbf{0.002}  \\
& \textbf{Ours (Scaffold-GS)} & \textbf{42.08}&	\textbf{0.981}&	\textbf{0.006}    &\textbf{45.40} &\textbf{0.993}	&\textbf{0.002}  &\textbf{37.65}&\textbf{0.971}&	\textbf{0.012}& \textbf{45.21}	&\textbf{0.994}	&\textbf{0.002}     \\    
\midrule     
 \multirow{5}[2]{*}{LDR-NE} & HDR-NeRF~\cite{hdr-nerf} &33.44&0.926&0.046&41.19&0.987&0.012&30.78&0.886&0.083&37.99&0.979&0.013       \\
% & HDR-NeRF~\cite{hdr-nerf} $\dag$ &  36.30     &  0.945  & 0.039 & 42.13& 0.989& 0.009   &  34.10& 0.929 &0.064 &39.81 &0.983&0.011  \\
& HDR-GS~\cite{hdr-gs} $\dag$ &  36.03	&0.963 &	0.024  & 41.69	&0.988	&0.005 &34.05 & 0.951	&0.026&34.62&	0.976&	0.020 \\
& \textbf{Ours (3DGS)} &  41.10&	0.977	&0.008&43.85 &0.992&	0.003 & 36.53&0.967	&0.015&38.78&0.980&0.021 \\
& HDR-Scaffold-GS~\cite{scaf-gs} $*$ &  40.61	&0.979	&0.008 &42.10	&0.992	&0.004 & 35.89&0.965&	0.017 & 41.00	&\textbf{0.993}	&\textbf{0.003}  \\
& \textbf{Ours (Scaffold-GS)} & \textbf{41.71} &	\textbf{0.982}&	\textbf{0.007}&\textbf{44.43} & \textbf{0.993}	&\textbf{0.002} &\textbf{37.14}&	\textbf{0.970}	&\textbf{0.013} &\textbf{43.33}&	\textbf{0.993}&	\textbf{0.003} \\    
\midrule
 \multirow{5}[2]{*}{HDR} & HDR-NeRF~\cite{hdr-nerf} &33.97&0.925&0.048&\textbf{43.22}&\textbf{0.991}&0.008&34.14&0.924&0.069&38.57&0.981&0.015     \\
% & HDR-NeRF~\cite{hdr-nerf} $\dag$ &  33.80& 0.921&0.050& 42.66 &0.990&0.008  & 31.82& 0.920&0.083     \\
& HDR-GS~\cite{hdr-gs} $\dag$ &  15.48	&0.739&	0.125 &30.99	&0.961 &0.027&20.47	&0.751	&0.115 &18.86&0.860&0.074 \\
& \textbf{Ours (3DGS)} &  34.87	&0.943&	0.024 &41.47&	0.987	&0.007 &36.75&0.961	&0.019 & 34.82	&0.967	&0.028 \\
& HDR-Scaffold-GS~\cite{scaf-gs} $*$ &  22.28	&0.873	&0.061  &29.79	&0.970&	0.021&24.27&0.842&	0.163&26.60&0.952	&0.037  \\
& \textbf{Ours (Scaffold-GS)} &\textbf{36.22}	&\textbf{0.952}&	\textbf{0.017} & 42.23 & 0.988&	\textbf{0.005}& \textbf{37.79}&	\textbf{0.966}	&\textbf{0.017}&\textbf{38.61}&\textbf{0.983}	&\textbf{0.007}\\  
    \bottomrule
    \end{tabular}}
       \begin{tablenotes}
        \footnotesize
        \item $\dag$ We re-implement HDR-GS~\cite{hdr-gs} under Exp-1 setting for fair comparison.
        \item $*$ We replace the scene representation in HDR-GS from 3DGS~\cite{3dgs} to Scaffold-GS~\cite{scaf-gs} to establish a baseline for our method utilizing Scaffold-GS.
     \end{tablenotes}
  \label{tab:hdrnerf-syn-8-1}%
\end{table*}%

\begin{table*}[htbp]
  \centering
  \caption{Per-scene quantitative comparisons on HDR-NeRF~\cite{hdr-nerf} synthetic datasets (Part 2). LDR-OE and LDR-NE denote the LDR results with exposure $\{t_1,t_3,t_5\}$ and $\{t_2,t_4\}$, respectively. HDR denotes the HDR results. The training exposure setting is Exp-1.}
  \resizebox{0.99\textwidth}{!}{
    \begin{tabular}{c|c|cccccccccccc}
    \toprule
&\multirow{2}[3]{*}{Method}  & \multicolumn{3}{c}{Dog} & \multicolumn{3}{c}{Desk} & \multicolumn{3}{c}{Sofa} & \multicolumn{3}{c}{Sponza} \\
\cmidrule(lr){3-5}\cmidrule(lr){6-8}\cmidrule(lr){9-11}\cmidrule(lr){12-14}    
&       & PSNR$\uparrow$  & SSIM$\uparrow$  & LPIPS$\downarrow$ & PSNR$\uparrow$  & SSIM$\uparrow$  & LPIPS$\downarrow$ & PSNR$\uparrow$  & SSIM$\uparrow$  & LPIPS$\downarrow$ & PSNR$\uparrow$  & SSIM$\uparrow$  & LPIPS$\downarrow$ \\
\midrule  
 \multirow{5}[2]{*}{LDR-OE} & HDR-NeRF~\cite{hdr-nerf} &37.77&0.981&0.016&  37.84&0.972&0.023 & 38.29&0.977&0.014 &  34.49&0.958&0.034     \\
% & HDR-NeRF~\cite{hdr-nerf} $\dag$ &       &       &       &       &       &       &       &       &       &       &       &        \\
& HDR-GS~\cite{hdr-gs} $\dag$ & 42.28 &	0.991	&0.004&39.81 &0.980	&0.007 & 41.75&0.989&0.004&37.99&0.972&0.016    \\
& \textbf{Ours (3DGS)} &43.06&	0.992&	0.004&42.28 &0.987&	0.004&43.04&0.990&\textbf{0.003}&41.43&0.986&0.007\\
& HDR-Scaffold-GS~\cite{scaf-gs} $*$ &  43.31&	0.993&	0.003 & 41.72 &	0.987&0.004 &43.26&\textbf{0.992}&\textbf{0.003} &42.43&0.990&\textbf{0.004} \\
& \textbf{Ours (Scaffold-GS)} & \textbf{43.76} &	\textbf{0.994}	&\textbf{0.002}   &\textbf{43.04}&	\textbf{0.990} &	\textbf{0.003} &\textbf{43.44} &	\textbf{0.992}	&\textbf{0.003} &\textbf{42.91}	&\textbf{0.991}&\textbf{0.004}  \\    
\midrule     
 \multirow{5}[2]{*}{LDR-NE} & HDR-NeRF~\cite{hdr-nerf} &36.52&0.976&0.018&35.26&0.960&0.029&38.35&0.976&0.014 &33.41&0.950&0.038 \\
% & HDR-NeRF~\cite{hdr-nerf} $\dag$ &       &       &       &       &       &       &       &       &       &       &       &        \\
& HDR-GS~\cite{hdr-gs} $\dag$ & 40.66 &	0.990&0.004  &38.47&	0.980&	0.007&41.55 & 0.989	&0.004 &36.99&0.975&0.015  \\
& \textbf{Ours (3DGS)} &  \textbf{42.24} &	0.991&	0.004  & 42.12 &	0.987	&0.004 & \textbf{42.34}&0.991&\textbf{0.003}&41.25&0.988&0.006 \\
& HDR-Scaffold-GS~\cite{scaf-gs} $*$ &  40.52 &	0.992&	0.004 &40.46&0.987	&0.004 &41.91&0.991&0.004 &41.02&0.990&0.004 \\
& \textbf{Ours (Scaffold-GS)} &  42.12 &	\textbf{0.993}	& \textbf{0.003}  &\textbf{42.76}&	\textbf{0.989}	&\textbf{0.003} &42.03&\textbf{0.992}	&\textbf{0.003}&\textbf{42.60}&\textbf{0.992}&\textbf{0.003} \\    
\midrule
 \multirow{5}[2]{*}{HDR} & HDR-NeRF~\cite{hdr-nerf} &\textbf{37.72}&\textbf{0.980}&0.016&43.38&\textbf{0.993}&0.007&\textbf{39.05}&\textbf{0.976}&0.017 &32.33&0.939&0.049   \\
% & HDR-NeRF~\cite{hdr-nerf} $\dag$ &       &       &       &       &       &       &       &       &       &       &       &        \\
& HDR-GS~\cite{hdr-gs} $\dag$ & 23.02 &	0.926&	0.037&29.11&0.773&0.082&26.60&0.928	&0.046&16.12&0.782&0.095  \\
& \textbf{Ours (3DGS)} &   36.36 &	0.973	& 0.016 &\textbf{43.98} &\textbf{0.993}&	0.006 &36.70&0.966&0.016 &34.32&0.964&0.023 \\
& HDR-Scaffold-GS~\cite{scaf-gs} $*$ &  24.33&	0.934&	0.040 &31.40&0.929&	0.044 &28.42&0.947&0.032&21.81&0.870&0.095   \\
& \textbf{Ours (Scaffold-GS)} &  37.52 &	0.978&	\textbf{0.012}  &43.83&\textbf{0.993}&\textbf{0.005} &36.99&0.969&\textbf{0.013} &\textbf{35.64}&\textbf{0.972}&\textbf{0.014}  \\  
    \bottomrule
    \end{tabular}}
       \begin{tablenotes}
        \footnotesize
        \item $\dag$ We re-implement HDR-GS~\cite{hdr-gs} under Exp-1 setting for fair comparison.
        \item $*$ We replace the scene representation in HDR-GS from 3DGS~\cite{3dgs} to Scaffold-GS~\cite{scaf-gs} to establish a baseline for our method utilizing Scaffold-GS.
     \end{tablenotes}
  \label{tab:hdrnerf-syn-8-2}%
\end{table*}%

\begin{figure*}[t]
\centering
%   \fbox{\rule{0pt}{2in} \rule{0.9\linewidth}{0pt}}
\includegraphics[width=0.96\linewidth]{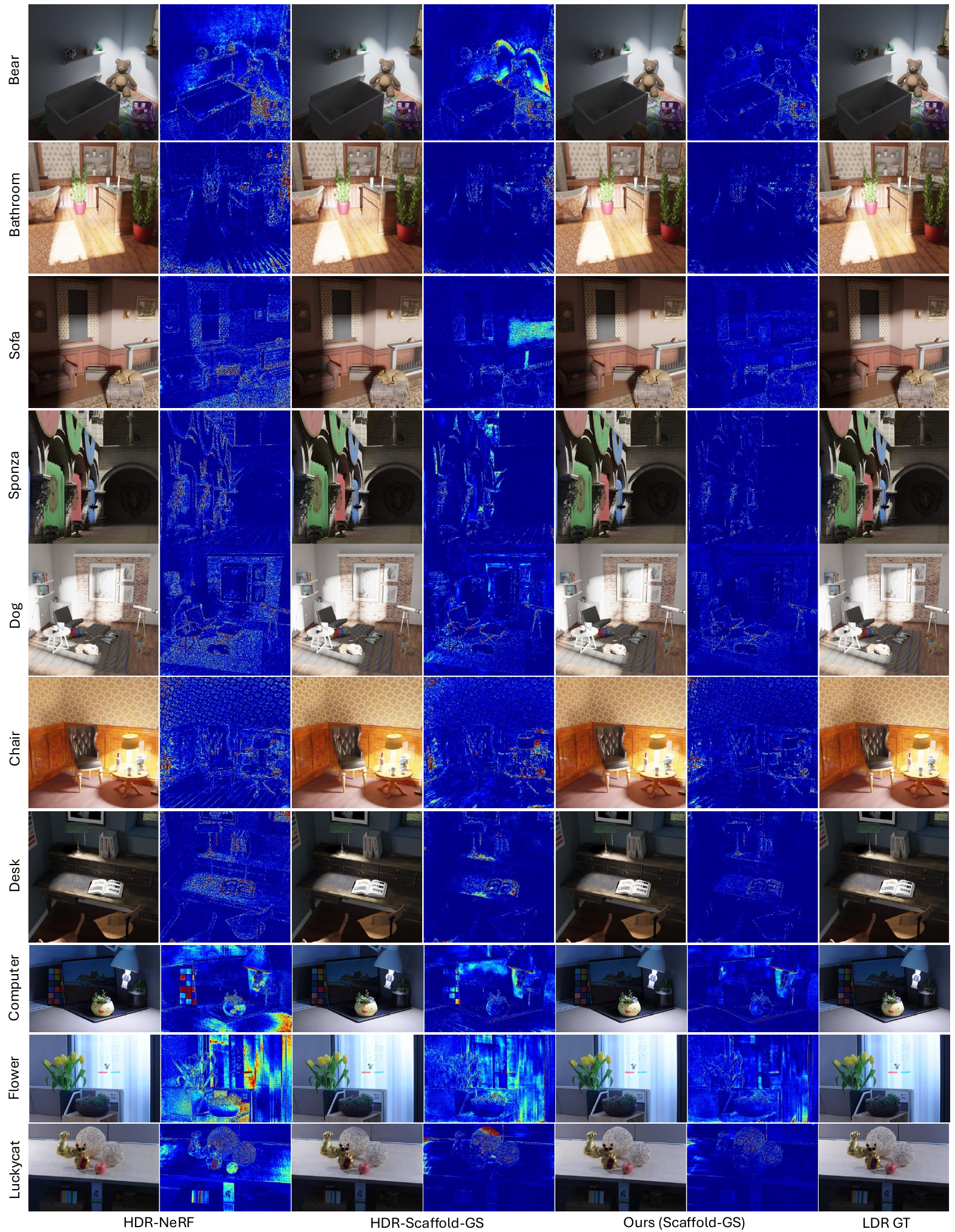}
\vspace{-3mm}
\caption{Qualitative LDR comparisons. Error maps in column 2, 4 and 6 show the MSE error compared to the ground truth, where color from blue to red indicates the error from small to large. Our method can reduce LDR fitting errors in some regions.} 
\label{fig:ldr_vis_sup}
\end{figure*}

\end{document}